\documentclass[runningheads]{llncs}

\usepackage[mobile]{eccv}

\usepackage{eccvabbrv}
\usepackage{paralist}
\usepackage{graphicx}
\usepackage{booktabs}

\usepackage[accsupp]{axessibility}  

\usepackage{multirow}
\usepackage{makecell}
\usepackage{wrapfig}
\newcommand{\modelName}{MALD-NeRF\xspace}

\newcommand{\degrade}[1]{{\color{red} #1}}
\newcommand{\improve}[1]{{\color{darkgreen} #1}}

\usepackage{color}
\definecolor{crimson}{rgb}{0.86, 0.08, 0.24}
\definecolor{gray}{rgb}{0.5,0.5,0.5}
\definecolor{green}{rgb}{0, 0.4, 0}
\definecolor{orange}{rgb}{1, 0.5, 0}
\definecolor{mahogany}{rgb}{0.75, 0.25, 0.0}
\definecolor{purple}{rgb}{0.6, 0, 0.6}
\definecolor{darkgreen}{rgb}{0, 0.4, 0}
\definecolor{frenchblue}{rgb}{0.0, 0.45, 0.73}
\definecolor{red}{rgb}{1,0,0}
\definecolor{yellow}{rgb}{1,1,0}
\definecolor{magenta}{rgb}{1,0,1}
\definecolor{pink}{rgb}{1,0.412,0.706}
\definecolor{newgreen}{rgb}{0, 0.6, 0.2}

\makeatletter
\DeclareRobustCommand\onedot{\futurelet\@let@token\@onedot}
\def\@onedot{\ifx\@let@token.\else.\null\fi\xspace}

\makeatother

\newlength\paramargin
\newlength\figmargin
\newlength\subfigmargin
\newlength\presecmargin
\newlength\secmargin
\newlength\subsecmargin
\newlength\tabmargin
\newlength\eqmargin
\newlength\paraskip

\long\def\ignorethis#1{}

\newcommand{\Paragraph}[1]
{\vspace{1mm} \noindent\textbf{#1}}

\makeatletter
\newcommand{\@chapapp}{\relax}%
\makeatother
\usepackage[title]{appendix}

\usepackage[pagebackref,breaklinks,colorlinks,citecolor=eccvblue,linkcolor=eccvblue]{hyperref}

\usepackage{orcidlink}

\begin{document}

\title{Taming Latent Diffusion Model for \protect\\ Neural Radiance Field Inpainting} 

\titlerunning{Taming Latent Diffusion Model for NeRF Inpainting}

\author{Chieh Hubert Lin\inst{1,2} \and
Changil Kim\inst{1} \and
Jia-Bin Huang\inst{1,3} \and
Qinbo Li\inst{1} \and
Chih Yao Ma\inst{1} \and
Johannes Kopf\inst{1} \and
Ming-Hsuan Yang\inst{2} \and
Hung-Yu Tseng\inst{1} \vspace{-.5em}
}

\authorrunning{C.~H.~Lin et al.}

\institute{$^1$Meta, $^2$University of California, Merced, $^3$University of Maryland, College Park
\\ [.5em]
\url{https://hubert0527.github.io/MALD-NeRF}}

\maketitle

\vspace{-1em}
\begin{figure}[h!]
    \centering
    \vspace{-.5em}
    \includegraphics[width=.95\linewidth]{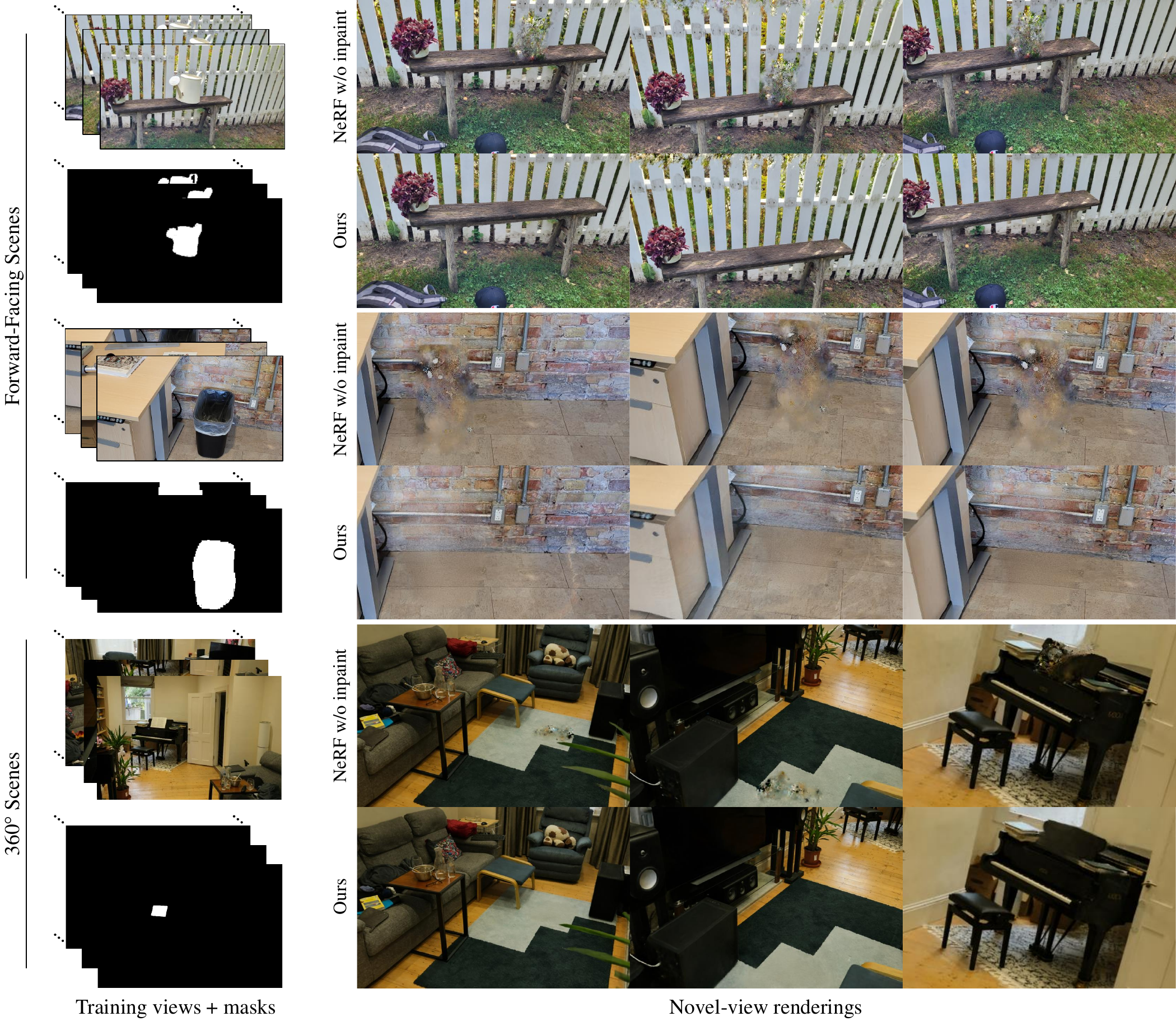}
    \vspace{-1em}
    \caption{\textbf{NeRF inpainting.} Given a set of posed images associated with inpainting masks, the proposed framework estimates a NeRF that renders high-quality novel views, where the inpainting region is realistic and contains high-frequency details. Our algorithm works for both forward-facing scenes and 360$^{\circ}$ scenes, and supports both single object and multiple objects removal. 
    }
    \label{fig:teaser}
    \vspace{-3.5em}
\end{figure}

\begin{abstract}

Neural Radiance Field (NeRF) is a representation for 3D reconstruction from multi-view images. 
Despite some recent work showing preliminary success in editing a reconstructed NeRF with diffusion prior, they remain struggling to synthesize reasonable geometry in completely uncovered regions.
One major reason is the high diversity of synthetic contents from the diffusion model, which hinders the radiance field from converging to a crisp and deterministic geometry.
Moreover, applying latent diffusion models on real data often yields a textural shift incoherent to the image condition due to auto-encoding errors.
These two problems are further reinforced with the use of pixel-distance losses.
To address these issues, we propose tempering the diffusion model's stochasticity with per-scene customization and mitigating the textural shift with masked adversarial training.
During the analyses, we also found the commonly used pixel and perceptual losses are harmful in the NeRF inpainting task.
Through rigorous experiments, our framework yields state-of-the-art NeRF inpainting results on various real-world scenes.
%
%

\end{abstract}
\vspace{\secmargin}
\section{Introduction}
\vspace{\secmargin}






The recent advancements in neural radiance fields (NeRF)~\cite{nerf,muller2022instant,barron2023zip} have achieved high-quality 3D reconstruction and novel-view synthesis of scenes captured with a collection of images.
The success intrigues an increasing attention on manipulating NeRFs such as 3D scene stylization~\cite{wang2023nerf,chiang2022stylizing} and NeRF editing~\cite{haque2023instruct}.
In this work, we focus on the \emph{NeRF inpainting} problem.
As shown in Figure~\ref{fig:teaser}, given a set of images of a scene with the inpainting masks, our goal is to estimate a completed NeRF that renders high-quality images at novel viewpoints.
The NeRF inpainting task enables a variety of 3D content creation applications such as removing objects from a scene~\cite{spinnerf,wang2023inpaintnerf360}, completing non-observed part of the scene, and hallucinating contents in the designated regions.
%
%
%


To address the NeRF inpainting problem, existing algorithms first leverage a 2D generative prior to inpaint the input images, then optimize a NeRF using the inpainted images.
Several efforts~\cite{spinnerf,mirzaei2023reference} use the LaMa~\cite{suvorov2022resolution} model as the 2D inpainting prior.
Driven by the recent success of diffusion models~\cite{dhariwal2021diffusion,ldm,saharia2205imagen,dai2023emu,betker2023dalle3}, recent work~\cite{inpaint3d,wang2023inpaintnerf360} use the diffusion models to further enhance the fidelity.
Nevertheless, unrealistic visual appearance and incorrect geometry are still observed in the inpainted NeRFs produced by these methods.

\begin{figure}[t]
    \centering
    \includegraphics[width=\textwidth]{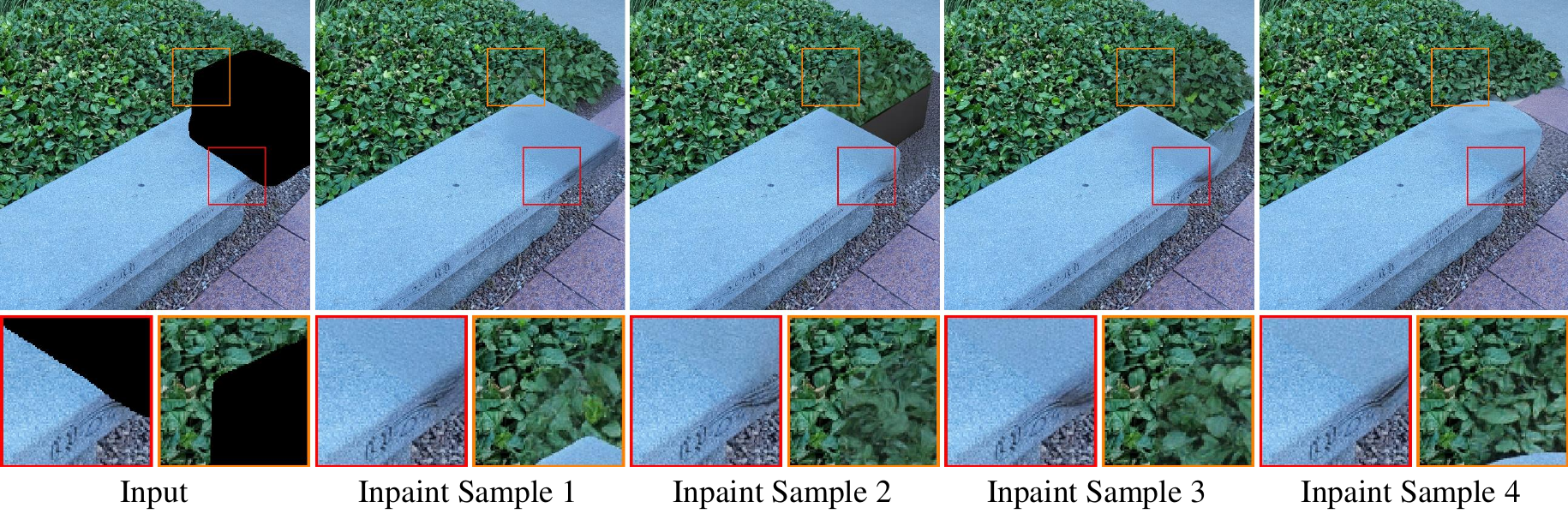}
    \vspace{-1.5em}
    \caption{\textbf{Inconsistency and texture shift issue.} We present the 2D inpainting results from our latent diffusion model. Given the same input image and mask, the results are 1) not consistent and 2) produce a texture shift between the original and inpainted pixels.
    These issues introduce noticeable artifacts in the NeRF inpainting results.
    }
    \label{fig:discontinuity}
    \vspace{-1em}
\end{figure}

Leveraging 2D latent diffusion models for NeRF inpainting is challenging for two reasons.
First, the input images inpainted by the 2D latent diffusion model are not 3D consistent.
The issue leads to blurry and mist-alike results in the inpainting region if pixel-level objectives (i.e., L1, L2) are used during NeRF optimization.
Several methods~\cite{spinnerf,wang2023inpaintnerf360} propose to use the perceptual loss function~\cite{lpips} to mitigate the issue.
Although the strategy improves the quality, the results still lack high-frequency details.
Second, as shown in Figure~\ref{fig:discontinuity}, the pixels inpainted by the latent diffusion model typically showcase a texture shift compared to the observed pixels in the input image.
The issue is due to the auto-encoding error in the latent diffusion model. 
It introduces noticeable artifacts in the final inpainted NeRF (i.e., the clearly visible seam between the reconstructed and inpainted region).

In this paper, we propose to use a masked adversarial training to address the two above-mentioned issues.
Our goal is to use the latent diffusion model to inpaint input images, and optimize a NeRF that 1) contains high-frequency details in the inpainted region, and 2) does not show texture difference between the inpainted and reconstructed regions.
Specifically, we introduce a patch-based adversarial objective between the inpainted and NeRF-rendered images to NeRF optimization.
Since the objective is not affixed to particular pixel similarity, we are capable of promoting high-frequency details without relying on absolutely consistent image pixels across the inpainted images.
However, simply applying the patch-based adversarial loss does not eliminate texture shift around the inpainting boundary, as it exists in the ``real'' examples (i.e., inpainted images) of the adversarial training.
To handle this, we design a masked adversarial training scheme to hide such boundaries from the discriminator, hence achieving improved quality.
In addition to the masked adversarial training, we apply per-scene customization to finetune the latent diffusion model~\cite{ruiz2023dreambooth,hu2021lora}, encouraging the model to generate contents that are more coherent to the reconstructed scene.
We find that the approach enhances the consistency across different input images, thus improving the quality of the final inpainted NeRF.

We conduct extensive quantitative and qualitative experiments on two NeRF inpainting benchmark datasets consisting of multiple challenging real-world scenes.
Our proposed method, name \modelName, achieves state-of-the-art NeRF inpainting performance by marrying the merits of the \underline{M}asked \underline{A}dversarial learning and the \underline{L}atent \underline{D}iffusion model.
\modelName synthesizes inpainting areas with high-frequency details and mitigates the texture shift issues.
In addition, we conduct extensive ablation studies to dissect the effectiveness of each component and loss function designs.
We summarize the contributions as follows:
\begin{compactitem}
    \item We design a masked adversarial training scheme for NeRF inpainting with diffusion. We show that the design is more robust to 3D and textural inconsistency caused by the 2D inpainting latent diffusion model.
    \item We harness the generation diversity of the latent diffusion model with per-scene customization. In combination with the iterative dataset update scheme, our framework yields better convergence while training inpainted NeRF.
    \item We achieve state-of-the-art NeRF inpainting performance.
\end{compactitem}


%

\begin{figure}[t]
    \centering
    \includegraphics[width=\textwidth]{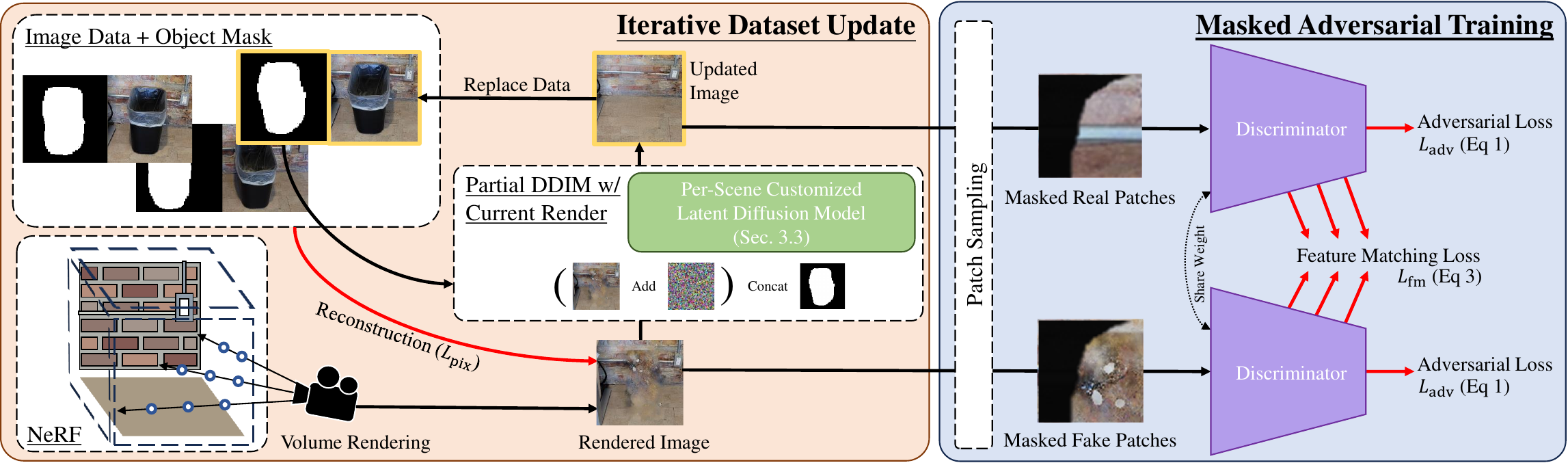}
    \vspace{-1.5em}
    \caption{
    \textbf{Method overview.}
    The proposed method uses a latent diffusion model to obtain the inpainted training images from the NeRF-rendered images using partial DDIM.
    The inpainted images are used to update the NeRF training dataset following the iterative dataset update protocol.
    (\textit{reconstruction}) We use pixel-level regression loss between the NeRF-rendered and ground-truth pixels to reconstruct the regions observed in the input images.
    (\textit{inpainting}) We design a masked patch-based adversarial training, which include an adversarial loss and discriminator feature matching loss, to supervise the the inpainting regions.
    %
    %
    %
    %
    }
    \label{fig:pipeline}
\end{figure}

\vspace{\secmargin}
\section{Related Work}
\vspace{\secmargin}

\vspace{\subsecmargin}
\subsection{3D Inpainting}
\vspace{\subsecmargin}
Inpainting 3D data is a long-standing problem in computer vision, such as point clouds~\cite{zhou2021pcd} and voxel~\cite{zhou2021pcd} completion. 
The classical approach is to collect a large-scale dataset of the target data distribution, then train a closed-form model on the data distribution by sampling random inpainting masks~\cite{suvorov2022resolution}.
However, to achieve high-quality results with a generalizable inpainting model, the approach requires a large-scale dataset that maintains high diversity and aligns with the testing distribution.
Currently, there is no existing large-scale NeRF dataset or feasible methods to directly train a closed-form inpainting model on NeRF representation.

\vspace{\subsecmargin}
\subsection{NeRF Inpainting}
\vspace{\subsecmargin}
Recent studies on NeRF inpainting focus on inpainting individual 2D images and mitigating the 3D inconsistency problem with different solutions. 
NeRF-In~\cite{shen2023nerfin} shows that simply training the NeRF with pixel reconstruction loss with inpainted images leads to blurry results.
SPIn-NeRF~\cite{spinnerf} is a more recent work that proposes that replacing pixel loss with more relaxed perceptual loss enables the NeRF to reveal more high-frequency details, and the method also leverages depth predictions to supervise the NeRF geometry.
\cite{weder2023removing} is a concurrent work of SPIn-NeRF, which also utilizes LaMa~\cite{suvorov2022resolution} as the image inpainter. 
%
%
InpaintNeRF360~\cite{wang2023inpaintnerf360} also adopts a similar strategy that relies on perceptual and depth losses.
However, in our study, we found using perceptual loss does not fundamentally resolve the problem, and often leads to sub-optimal quality.
Reference-guided inpainting~\cite{mirzaei2023reference} introduces a more sophisticated mechanism with careful per-view inpainting, view selection, and inpainting by referencing previous results to address the cross-view inconsistency.
However, the approach is highly heuristic and sophisticated, the authors have not released either the implementation or rendering results.
We have reached out to the authors to obtain visual results for comparison, but we did not receive responses.
Inpaint3D~\cite{inpaint3d} proposes to train a diffusion model on the RealEstate10k~\cite{realestate} dataset and shows high-quality scenes from a similar data distribution.
However, due to the limited size of the training dataset, the method is unable to generalize to arbitrary scenes, such as the SPIn-NeRF benchmark.
In contrast, we utilize a diffusion model pretrained on a large-scale internal dataset that serves as a strong prior model with outstanding generalization and shows that our method can be applied to unseen dataset distribution, such as the SPIn-NeRF benchmark.
NeRFiller~\cite{weber2023nerfiller} is a concurrent work that also works updating NeRF with constrained inpainting results using diffusion prior.

\vspace{\subsecmargin}
\subsection{Sparse-View NeRF Reconstruction Using Generative Prior}
\vspace{\subsecmargin}

Reconstructing a NeRF from a sparse set of a few images~\cite{yu2021pixelnerf} is a popular research topic due to its wide applications.
Several recent work proposes to utilize generative priors, such as diffusion models~\cite{wynn2023diffusionerf,liu2023deceptive,wu2023reconfusion} and GANs~\cite{roessle2023ganerf}.
Despite utilizing generative priors, the problem focuses on finding reasonable geometric correspondence from ill-conditioned sparse image sets and enhancing the surface quality of low-coverage regions.
The line of work does not consider the visual plausibility of uncovered regions, nor intentionally create disocclusion by removing objects.
It is worth noting such a distinction is significant, as the sparse-view reconstruction problem assumes the true geometry is accessible (despite being ill-conditioned) from the training views with fully trusted pixels.
Therefore, these algorithms are less concerned with the cross-view inconsistency problems caused by the generative models.
In contrast, the NeRF inpainting problem requires the algorithms to form the inpainted geometry from scratch.

\vspace{\secmargin}
\section{Methodology}
\vspace{\secmargin}

\vspace{\subsecmargin}
\subsection{Preliminaries}
\vspace{\subsecmargin}

\Paragraph{Neural radiance fields (NeRFs).} Given a set of $N$ images $\{I_i\}_{i=1 \dots N}$ with camera poses $\{P_i\}_{i=1\dots N}$, NeRFs aim to represent the 3D scene using a neural function $f_\theta$.
The neural function $f_\theta$, which can be implicit MLPs~\cite{nerf} or voxelized 3D volumes~\cite{muller2022instant}, learns to map the 3D position along with viewing direction to the corresponding density and color.
By applying volume rendering~\cite{drebin1988volume,nerf}, we can optimize a NeRF using a pixel-level regression loss between the rendered pixels $\{x\}$ and ground-truth pixels $\{\hat{x}\}$.

\Paragraph{NeRF inpainting.} In addition to the images, we are given a set of binary masks $\{M_i\}_{n=1\dots N}$ in this problem.
The binary masks split the image pixels into two distinct sets: the unmasked pixels $\{x^r_j\}$ that are used for reconstructing the observed part of the scene, and the masked pixels $\{x^m\}$ indicating the unknown regions to be inpainted.

\vspace{\subsecmargin}
\subsection{NeRF Inpainting with Latent Diffusion Models}
\vspace{\subsecmargin}

Figure~\ref{fig:pipeline} presents an overview of the proposed framework.
We use a latent diffusion model pre-trained on an internal image inpainting dataset to inpaint the 2D images, then replace the input images for NeRF training.
The pixel-level loss function is used for reconstructing the known region in the input images, while the masked adversarial training is used for the inpainting region. 
We detail each component as follows.

\Paragraph{Reconstructing observed regions.} For reconstruction pixels, we supervise the NeRF model with an L2 objective $L_\text{pix} = \| x^r_j - \hat{x}^r_j \|_2$.
The inter-level loss $L_\text{inter}$~\cite{barron2021mip}, distortion loss $L_\text{distort}$~\cite{barron2021mip} and hash decay $L_\text{decay}$~\cite{barron2023zip} are also used for regularization.

\Paragraph{Masked adversarial training for inpainting regions.}
We do not use pixel distance losses in the inpainting region, as they are not robust to the highly diverse and 3D inconsistent inpainting results leading to blurry mist-like NeRF renderings.
To address the issue, we use adversarial loss~\cite{goodfellow2014generative,stylegan2} and the discriminator feature matching loss~\cite{wang2018high} to guide the NeRF in the inpainting regions.
Specifically, we consider the patches of inpainted images as real examples, and the NeRF-rendered patches as the fake ones in the adversarial training.

However, as shown in Figure~\ref{fig:discontinuity}, the real pixels and the inpainted pixels have a textural shift, causing the discriminator to exploit such a property to recognize the real image patches.
In this case, the discriminator promotes the textural discrepancy between the NeRF-rendered pixels in the reconstruction and inpainting regions.
To alleviate the issue, we design a masked adversarial training scheme to hide the reconstruction/inpainting boundary on the image patches from the discriminator.
For both NeRF-rendered and diffusion-inpainted images, we only keep the pixels within the inpainting mask region, and mask the pixels outside the masked region with black pixels.
%
%
The design is conceptually similar to AmbientGAN~\cite{bora2018ambientgan} which trains the discriminator with corrupted inputs based on the underlying task of interest.
In particular, our objective is to reduce the textural gap between the inpainting and non-inpainting renderings, instead of solving image restoration tasks.
We later show the design indeed improves inpainting geometry and performs better in quantitative evaluations in the ablation study.

Given the masking functions $C^m$ for the inpainting region and $C^r$ for the non-inpainting region, the adversarial loss for the discriminator $D$ training is
\begin{equation}
L_\text{adv} = f(D(C^m(x^m))) + f(-D(C^r(\hat{x}^r)))
\end{equation}
where $f(x) = - \text{log}(1+\text{exp}(-x))$.
We use the StyleGAN2~\cite{stylegan2} discriminator architecture and train the discriminator with patches~\cite{isola2017image}.
In addition, we use R1 regularizer~\cite{mescheder2018training} to stabilize the discriminator training with
\begin{equation}
L_\text{GP} = \| \nabla D(C^r(\hat{x}^r)) \|^2_2 \, .
\end{equation}
Meanwhile, we extract the discriminator intermediate features after each discriminator residual blocks with $F$, then calculate the discriminator feature matching loss~\cite{wang2018high} to supervise the inpainting area
\begin{equation}
L_\text{fm} = \| F(C^m(x^m)) - F(C^m(\hat{x}^m) \|_1 \, .
\end{equation}

\Paragraph{Monocular depth supervision for inpainting regions.}
%
We leverage an off-the-shelf monocular depth prior to regularize the geometry of the learned NeRF. 
We use ZoeDepth~\cite{bhat2023zoedepth} to estimate the depth $\tilde{d}_i$ of the inpainted images, then render the NeRF depth $d_i$ by integrating the density along the radiances.
Since the two depths are in different metrics, similar to \cite{hollein2023text2room}, we solve a shift-scale factor between the two depth maps.
In particular, we only use the reconstruction region to compute the shift-scale factor, and use the solved factor to rescale the estimated depth into the final depth supervision $\hat{d}_i$.
Since the computation requires meaningful NeRF depths, we start applying the depth supervision after 2{,}000 iterations.
We use a ranking depth loss $L_D$ proposed in SparseNeRF~\cite{wang2023sparsenerf}.

\vspace{\paraskip}
\noindent \textbf{Total training objective.}
Each training iteration of \modelName consists of three steps, each step optimizes the modules with different objectives.
A reconstruction step optimizing NeRF with
$
    L^r = \lambda_\text{pix} L_\text{pix} + \lambda_\text{inter} L_\text{inter} + \lambda_\text{distort} L_\text{distort} + \lambda_\text{decay} L_\text{decay} \, .
$
%
An inpainting step optimizing NeRF with
$
L^m = - \lambda_\text{adv} L_\text{adv} + \lambda_{fm} L_\text{fm} + \lambda_\text{inter} L_\text{inter} + \lambda_\text{distort} L_\text{distort} + \lambda_\text{decay} L_\text{decay} \, .
$
Finally, the discriminator training step using
$L^D = L_\text{adv} + \lambda_\text{GP} L_\text{GP}.$
The $\lambda$'s are objective weights.

\Paragraph{Iterative dataset update and noise scheduling in inpainting diffusion.}
In practice, directly inpainting the input images leads to inconsistencies across various viewpoints due to the high diversity and randomness of diffusion models.
We leverage two strategies to mitigate the issue.
First, we use an iterative dataset update (IDU) similar to \cite{haque2023instruct}, where we gradually update the inpainting region every $U$ iterations throughout the training.
Second, the inpainting diffusion model only performs a partial DDIM~\cite{song2020denoising} starting from time step $t$ based on the current NeRF rendering.
The time step $t$ is determined based on the ratio of the training progress, i.e., the earlier the training is, the more noise is added to the inpainting region for denoising.
Such a design aims to leverage the 3D consistency of NeRF rendering and gradually propagate the 3D consistent information to all images in the dataset.
Specifically, we use the HiFA~\cite{zhu2023hifa} scheduling that sets $t= t_{max} - (t_{max} - t_{min}) * \sqrt{k / K}$, where $k$ is the current NeRF training time step, $K$ is the total iterations of the NeRF training, and $(t_{max}, t_{min})=(980, 20)$ is the DDPM~\cite{ho2020denoising} time steps used within the latent diffusion model.

\vspace{\subsecmargin}
\subsection{Per-Scene Customized Latent Diffusion Model}
\vspace{\subsecmargin}

In order to harness the expressiveness of the diffusion model and avoid synthesizing too many out-of-context objects that confuse the convergence of the inpainted NeRF.
We finetune the inpainting diffusion model in each scene.
For each scene, we set a customized text token for the scene, and LoRA~\cite{hu2021lora} finetune both the text encoder and U-Net.
We use a self-supervised inpainting loss similar to \cite{suvorov2022resolution,tang2023realfill}.
For each image, we sample arbitrary rectangular inpainting masks and take the union of masks as the training mask.
The LoRA-finetuning model is being supervised to inpaint the latent values within the training mask.
Following the DDPM~\cite{ho2020denoising} training, we supervise the U-Net of the diffusion with L2 distance at a random time step.
%
Meanwhile, since we are working on an object removal task, each image is paired with a 3D consistent mask that does not have ground-truth supervision.
We set the loss values to zero within the object removal mask region.





\vspace{\secmargin}
\section{Experiments}
\vspace{\secmargin}

\vspace{\subsecmargin}
\subsection{Experimental Setups}
\vspace{\subsecmargin}

\Paragraph{Datasets.} We use two real-world datasets for all experiments:
\begin{compactitem}
\item \textbf{SPIn-NeRF}~\cite{spinnerf} is an object removal benchmark dataset consisting of $10$ scenes.
Each scene has $60$ training views captured with an object (to be removed), and each view is associated with an inpainting mask indicating the desired object to be removed.
In addition, each scene contains $40$ testing views in which the object is physically removed during the capture.
\item \textbf{LLFF}~\cite{llff} consists of multiple real-world scenes with varying numbers of images ($20$-$45$). 
We use a six-scene subset provided by SPIn-NeRF annotated with 3D grounded object removal masks.
\end{compactitem}
Following \cite{spinnerf,inpaint3d}, we resize all the images to have a long-edge size of $1008$.

%
%
%
%
%

\begin{figure}[t]
    \centering
    \includegraphics[width=\textwidth]{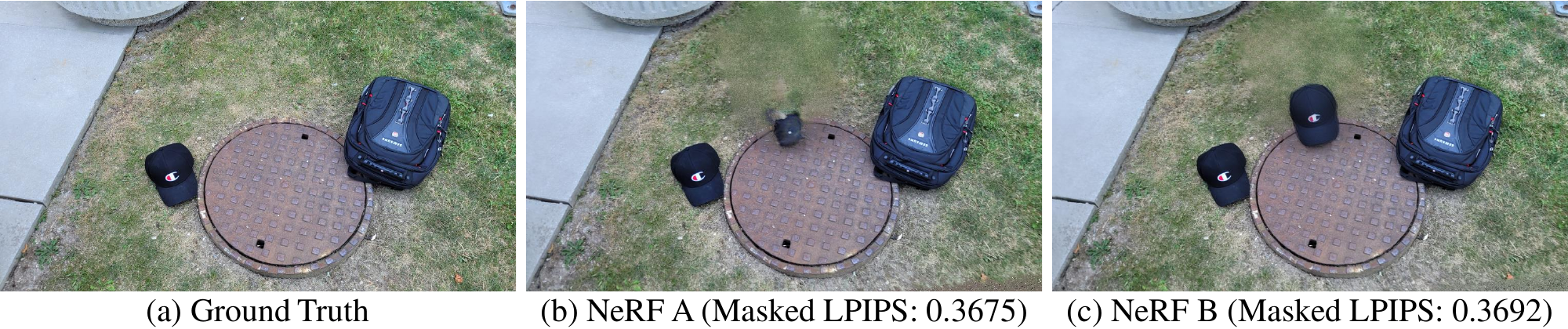}
    \vspace{-1.5em}
    \caption{\textbf{Drawbacks of LPIPS.} In some cases, the LPIPS score fails to indicate the visual quality.
    For example, generating a realistic baseball cap actually lowers the score as there is no object in the inpainting area in the ground truth image.}
    \label{fig:lpips-problem}
\end{figure}

\Paragraph{Evaluation setting.} We follow the protocal in the SPIn-NeRF~\cite{spinnerf} work.
We optimize the NeRF using the training view images (with objects) associated with the inpainting masks for all compared methods.
We only use the test views, where the object is physically removed from the scene, to compute the below metrics for evaluation:
\begin{compactitem}
\item \textbf{LPIPS}: We use the LPIPS score~\cite{lpips} to measure the perceptual difference between the NeRF-rendered and ground-truth test view images.

\item \textbf{M-LPIPS}: To better understand the inpainting performance, we \emph{mask out} the region outside the object inpainting mask and measure the LPIPS score.

\item \textbf{FID}/\textbf{KID}: As shown in Figure~\ref{fig:lpips-problem}, the LPIPS score is not a proper metric for generative tasks.
For instance, although the object generated in the inpainting area is valid content, it produces a high perceptual distance to the ground truth test view.
To address the issue, we additionally report the FID~\cite{fid} and KID~\cite{kid} scores, which are commonly used metrics in generative model literature that quantify the distributional similarity between two sets of images and are sensitive to the visual artifacts.
For each evaluated method, we compute the scores using NeRF-rendered and ground-truth images of all test views across all scenes in the dataset.
\item \textbf{C-FID}/\textbf{C-KID}: For the LLFF dataset, since the dataset does not include test views with the object being physically removed, we alternatively measure the visual quality near the inpainting border. More specifically, we find the four furthest corners of the inpainting mask and crop image patches centered at these corners. 
Then, finally, compute the FID/KID scores between the real-image patches and the NeRF-rendering after the object is removed and inpainted.
\item \textbf{FVD}: To quantify the geometric consistency, we consider the image set of each scene as a video sequence, then quantify the similarity between the real and inpainted sequences. We use FVD~\cite{unterthiner2019fvd} that measures both the perceptual and temporal consistency of video sequence. In order to create sufficient evaluation samples, we sub-sample each scene into multiple short sequences, each has a length of ten frames. However, the blurry results can have better temporal consistency and result in better FVD scores, therefore, the metric should be considered along with previously mentioned FID/KID.
\end{compactitem}

%
%
%
%
%

\begin{table}[t]
\centering
\setlength{\tabcolsep}{5pt}
\caption{\textbf{Quantitative comparisons.} We present the results on the SPIn-NeRF~\cite{spinnerf} and LLFF~\cite{llff} datasets. 
Note that the LLFF dataset does not have ground-truth views with object being physically removed, therefore, we only measures C-FID and C-KID on these scenes.
The best performance is \underline{underscored}.
}
\label{tab:quant}
\vspace{-.5em}
\scriptsize
\begin{tabular}{@{}lccccccc@{}}
\toprule
Methods & \multicolumn{4}{c}{SPIn-NeRF} & \multicolumn{2}{c}{LLFF} \\ \cmidrule(l){2-5} \cmidrule(l){6-7}
                         & LPIPS ($\downarrow$) & M-LPIPS ($\downarrow$) & FID ($\downarrow$) & KID ($\downarrow$) & C-FID ($\downarrow$) & C-KID ($\downarrow$) \\ \midrule
SPIn-NeRF                & 0.5356 & 0.4019 & 219.80  & 0.0616 & 231.91  & 0.0654 \\
SPIn-NeRF (LDM)          & 0.5568 & 0.4284 & 227.87  & 0.0558 & 235.67  & 0.0642 \\
Inpaint3d                & 0.5437 & 0.4374 & 271.66  & 0.0964 & $-$     & $-$    \\
InpaintNeRF360           & 0.4694 & 0.3672 & 222.12  & 0.0544 & 174.55  & 0.0397 \\
Ours                     & \underline{0.4345} & \underline{0.3344} & \underline{183.25}  & \underline{0.0397} & \underline{171.89}  & \underline{0.0388} \\ \bottomrule
\end{tabular}
\end{table}

\Paragraph{Evaluated methods.} We compare our method with the following approaches:
\begin{compactitem}
\item \textbf{SPIn-NeRF}~\cite{spinnerf}: 
Note that the authors do not provide the evaluation implementation.
Therefore, the LPIPS scores reported in this paper differ from those presented in the SPIn-NeRF paper.
Nevertheless, we have contacted the authors to ensure that the SPIn-NeRF results match the quality shown in the original paper.

\item \textbf{SPIn-NeRF (LDM)}: 
We use our latent diffusion model in the SPIn-NeRF approach as the inpainting module while maintaining all default hyper-parameter settings.

\item \textbf{InpaintNeRF360}~\cite{wang2023inpaintnerf360}: We implement the algorithm as no source code is available. 
Specifically, we use our latent diffusion model for per-view inpainting and optimize the NeRF with the same network architecture devised in our approach with the objectives proposed in the paper.

\item \textbf{Inpaint3d}~\cite{inpaint3d}: We reach out to the authors for all the rendered images of test views for evaluation on the SPIn-NeRF dataset.

\end{compactitem}






\begin{figure}[t!]
    \centering
    \setlength{\tabcolsep}{1pt}
    \begin{tabular}{c c cc cc}
        \toprule
        \multirow{2}{*}{
            \includegraphics[width=.18\linewidth,trim={0cm 0cm 0 0},clip]{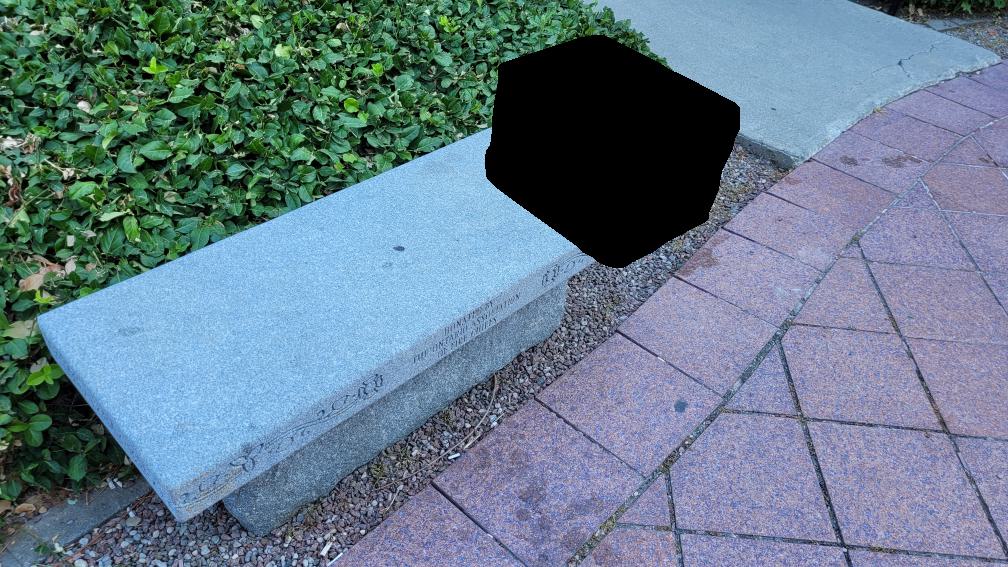}
        } &
        \parbox[t]{1em}{\rotatebox[origin=l]{90}{\makecell{\scalebox{.65}{ \hspace{0.0em} w/o Customization}}}}  &
        \includegraphics[width=.18\linewidth,trim={0cm 0cm 0 0},clip]{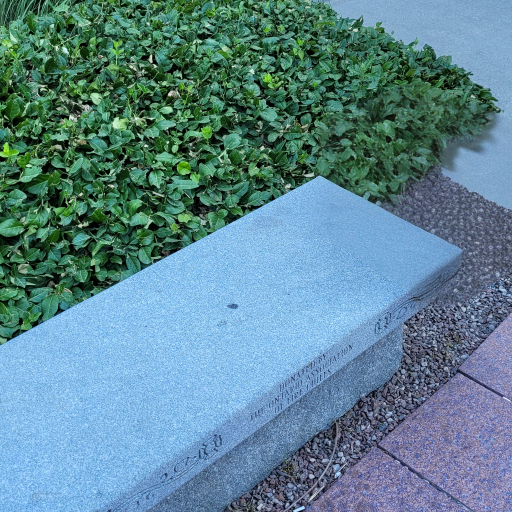} & 
        \includegraphics[width=.18\linewidth,trim={0cm 0cm 0 0},clip]{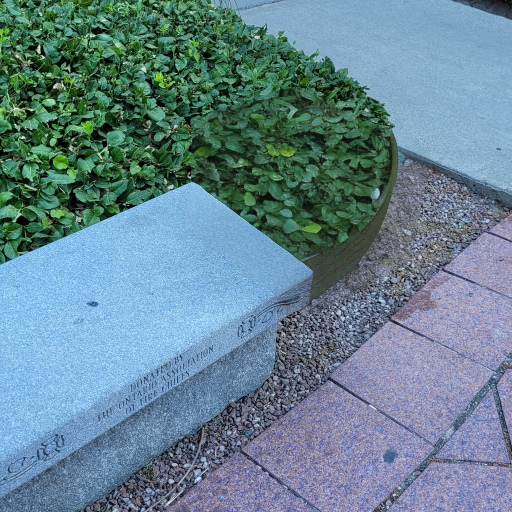} & 
        \includegraphics[width=.18\linewidth,trim={0cm 0cm 0 0},clip]{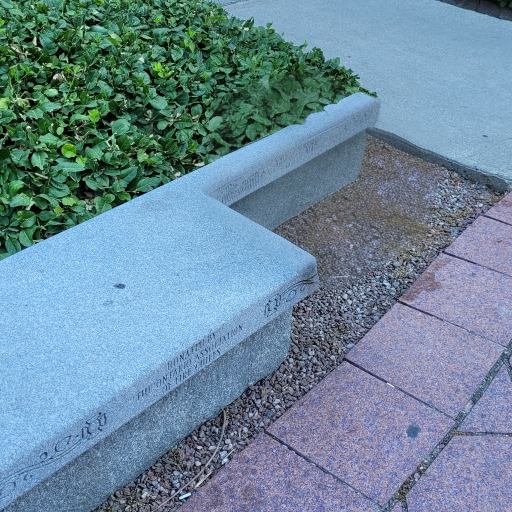} & 
        \includegraphics[width=.18\linewidth,trim={0cm 0cm 0 0},clip]{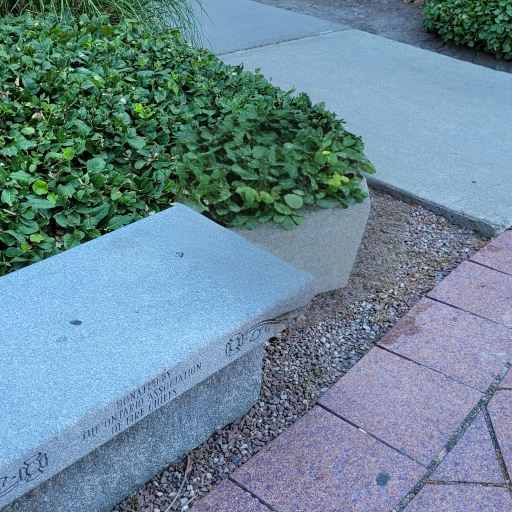} \\ [-0.3em]
        &
        \parbox[t]{1em}{\rotatebox[origin=l]{90}{\makecell{\scalebox{.65}{ \hspace{0.25em} w/ Customization}}}}  &
        \includegraphics[width=.18\linewidth,trim={0cm 0cm 0 0},clip]{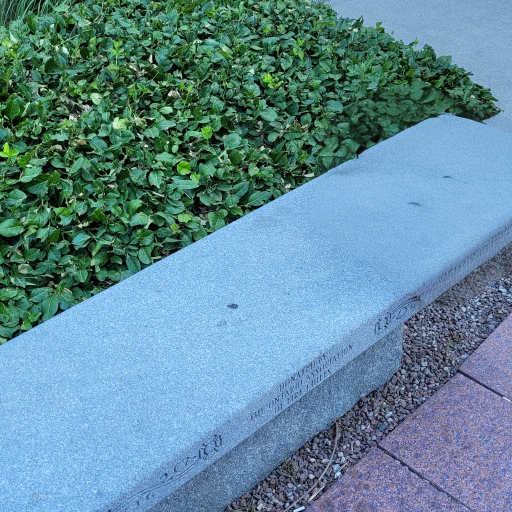} & 
        \includegraphics[width=.18\linewidth,trim={0cm 0cm 0 0},clip]{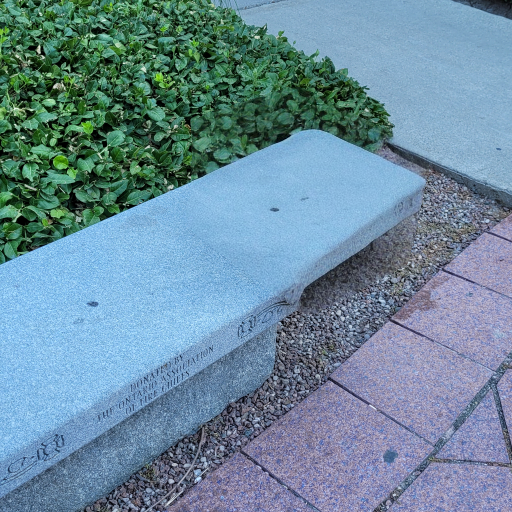} &
        \includegraphics[width=.18\linewidth,trim={0cm 0cm 0 0},clip]{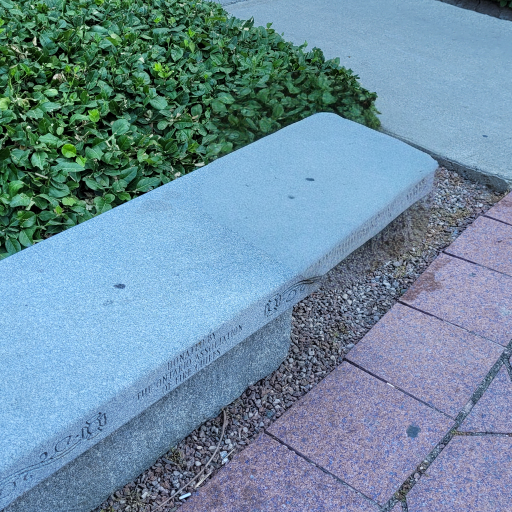} &
        \includegraphics[width=.18\linewidth,trim={0cm 0cm 0 0},clip]{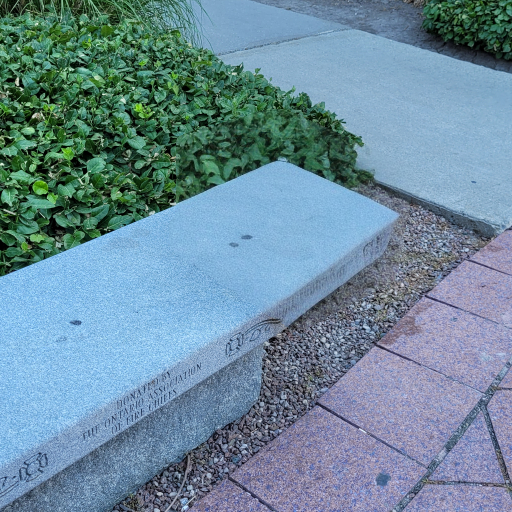} \\ [-0.3em]
        \midrule
        \multirow{2}{*}{
            \includegraphics[width=.18\linewidth,trim={0cm 0cm 0 0},clip]{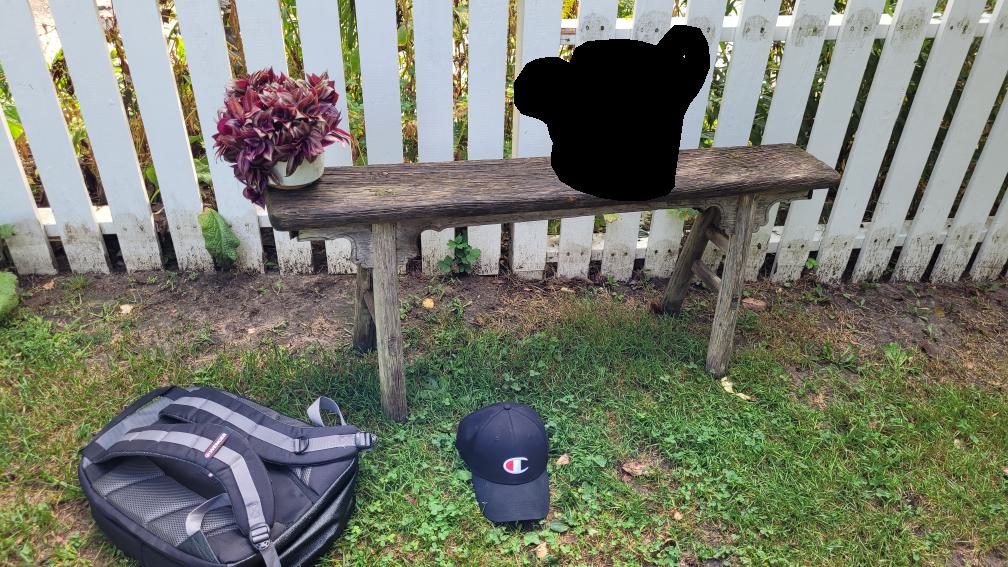}
        } &
        \parbox[t]{1em}{\rotatebox[origin=l]{90}{\makecell{\scalebox{.65}{ \hspace{0em} w/o Customization}}}} &
        \includegraphics[width=.18\linewidth,trim={0cm 0cm 0 0},clip]{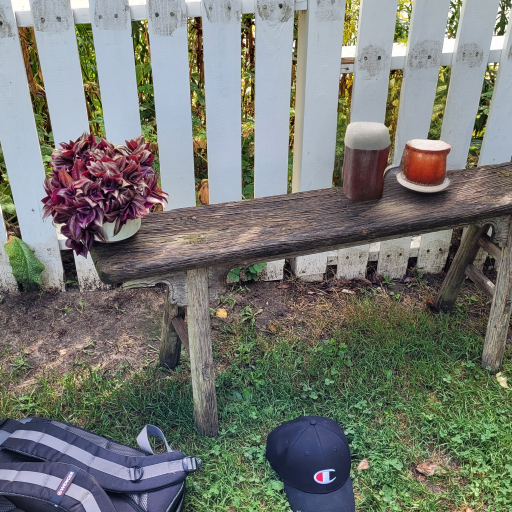} & 
        \includegraphics[width=.18\linewidth,trim={0cm 0cm 0 0},clip]{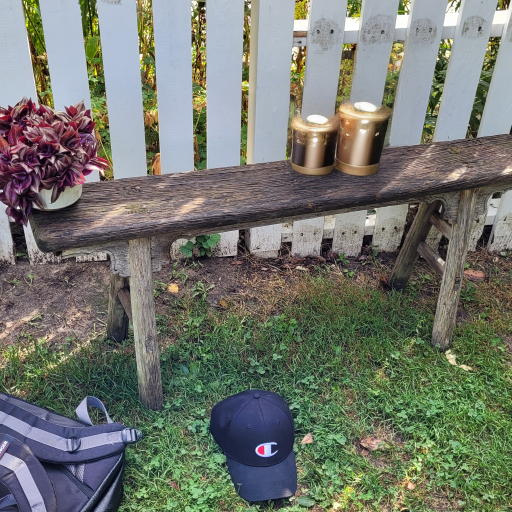} &
        \includegraphics[width=.18\linewidth,trim={0cm 0cm 0 0},clip]{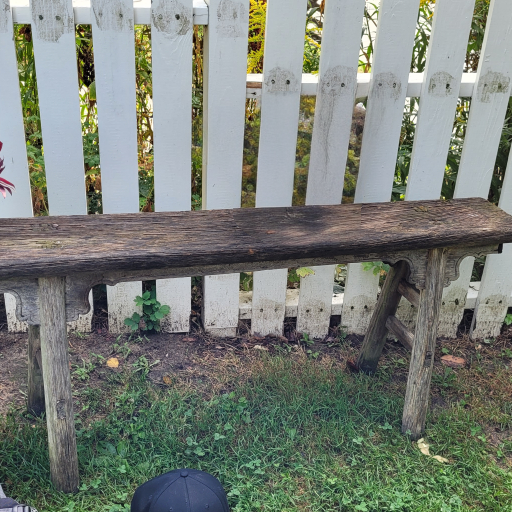} &
        \includegraphics[width=.18\linewidth,trim={0cm 0cm 0 0},clip]{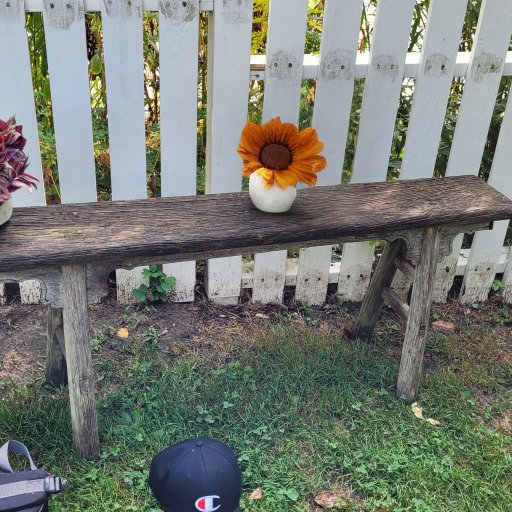}\\ [-0.3em]
        &
        \parbox[t]{1em}{\rotatebox[origin=l]{90}{\makecell{\scalebox{.65}{ \hspace{0.25em} w/ Customization}}}} &
        \includegraphics[width=.18\linewidth,trim={0cm 0cm 0 0},clip]{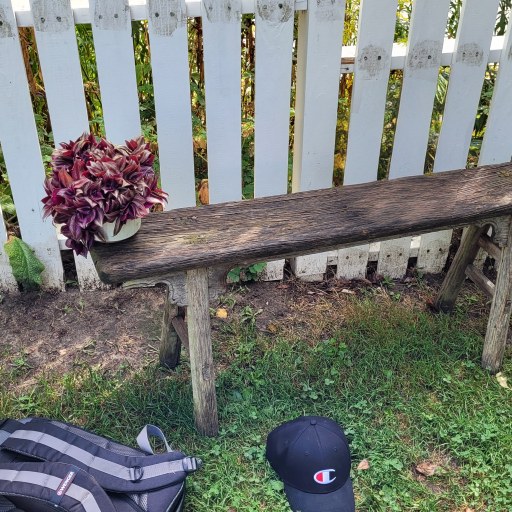} & 
        \includegraphics[width=.18\linewidth,trim={0cm 0cm 0 0},clip]{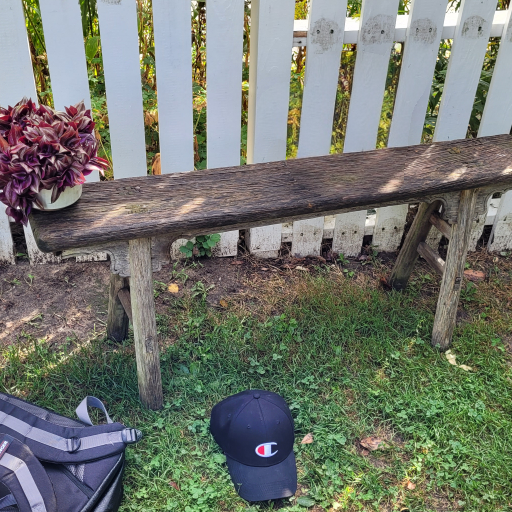} &
        \includegraphics[width=.18\linewidth,trim={0cm 0cm 0 0},clip]{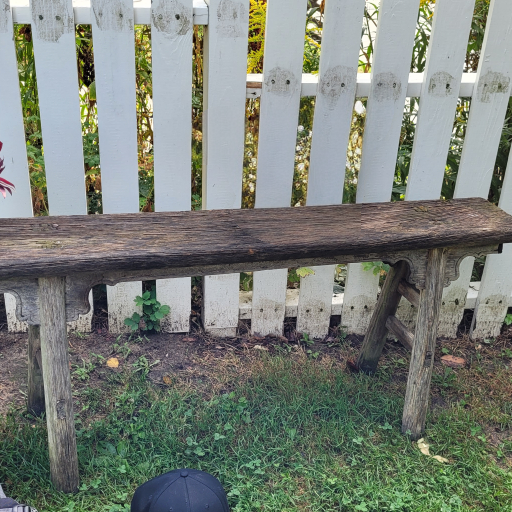} &
        \includegraphics[width=.18\linewidth,trim={0cm 0cm 0 0},clip]{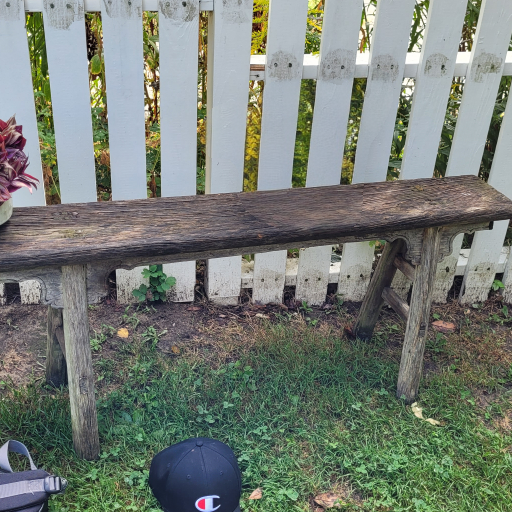} \\ [-.3em]
        \bottomrule
    \end{tabular}
    \vspace{-.5em}
    \caption{\textbf{Per-scene customization.} Our per-scene customization effectively forges the latent diffusion model to synthesize consistent and in-context contents across views.}
    \label{exp:ldm-finetune}
    \vspace{-.5em}
\end{figure}

\begin{figure}[t!]
    \centering
    \setlength{\tabcolsep}{0pt}
    \begin{tabular}{c cc cc}
        %
        \parbox[t]{1.2em}{\rotatebox[origin=l]{90}{\makecell{\scalebox{.55}{\hspace{.75em} Input + Mask}}}} &
        \includegraphics[width=.24\linewidth,trim={0 0 7.11cm 4cm},clip]{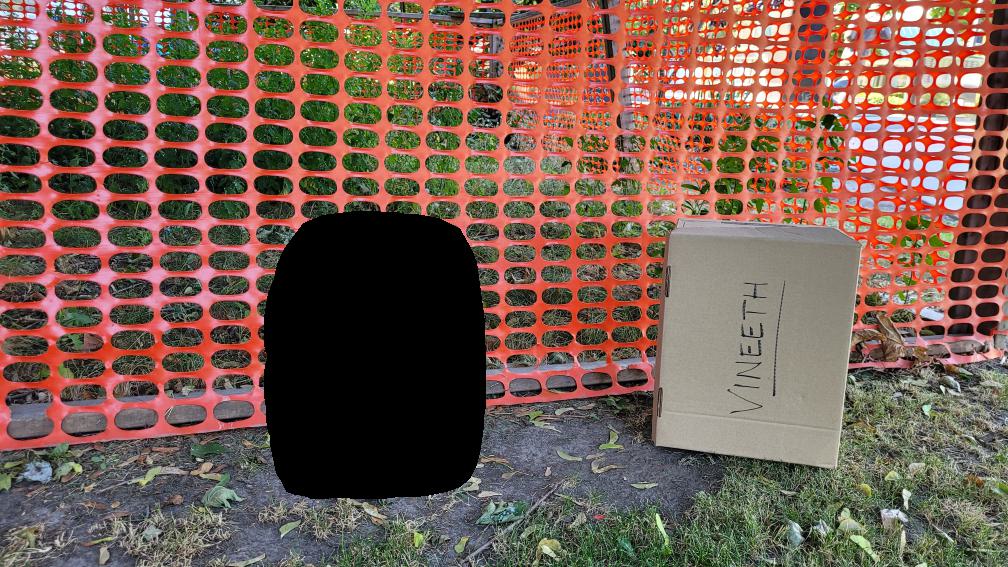} &
        \includegraphics[width=.24\linewidth,trim={0 0 7.11cm 4cm},clip]{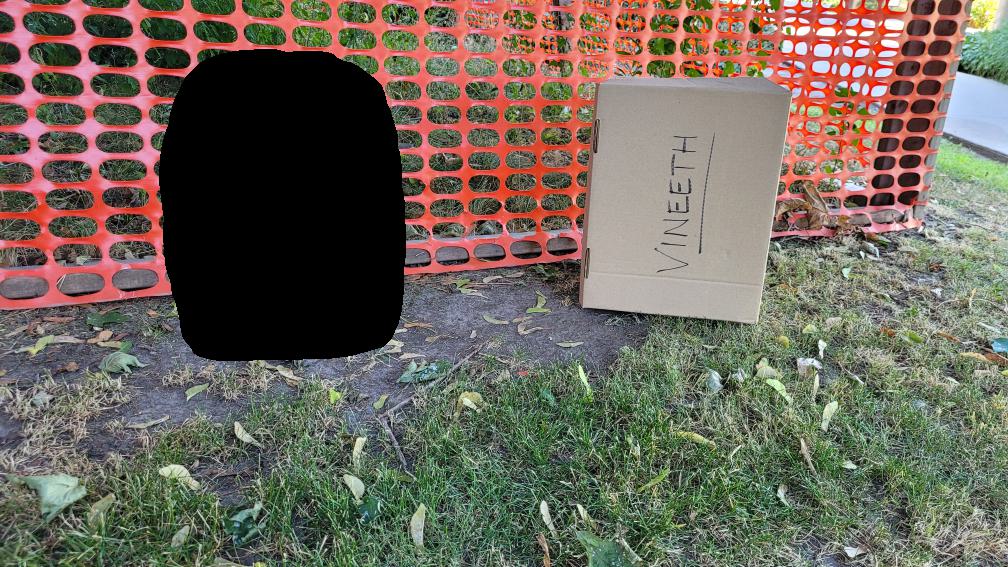} &
        \includegraphics[width=.24\linewidth,trim={0cm 0cm 0 0},clip]{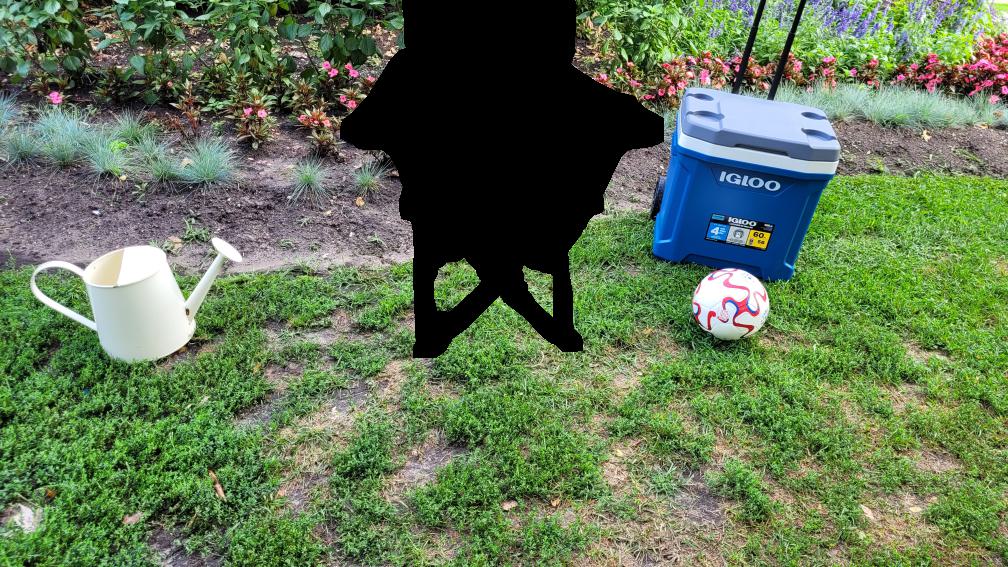} &
        \includegraphics[width=.24\linewidth,trim={0cm 0cm 0 0},clip]{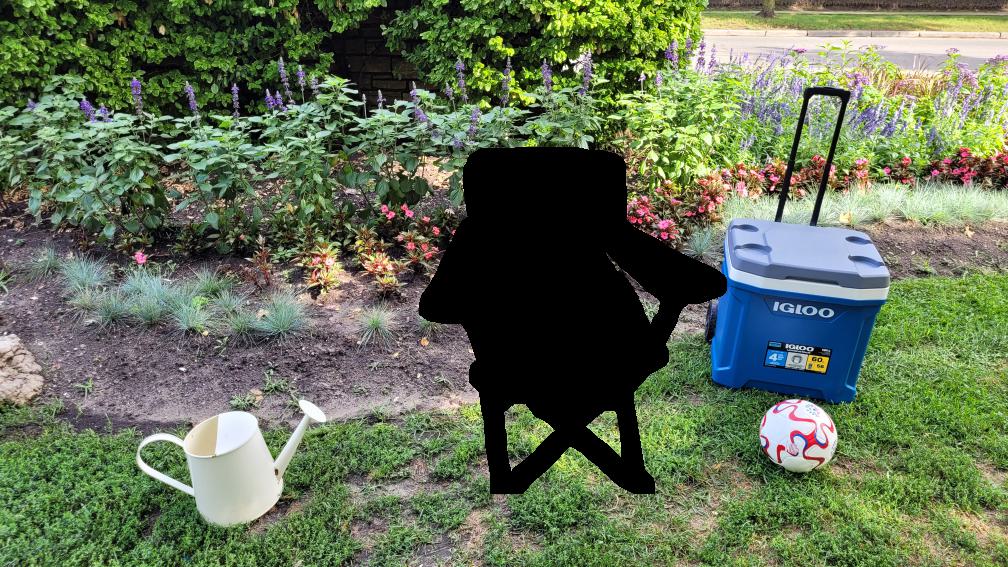} \\ [-0.25em]
        %
        \parbox[t]{1.2em}{\rotatebox[origin=l]{90}{\makecell{\scalebox{.55}{\hspace{1em} SPIn-NeRF}}}} &
        \includegraphics[width=.24\linewidth,trim={0 0 7.11cm 4cm},clip]{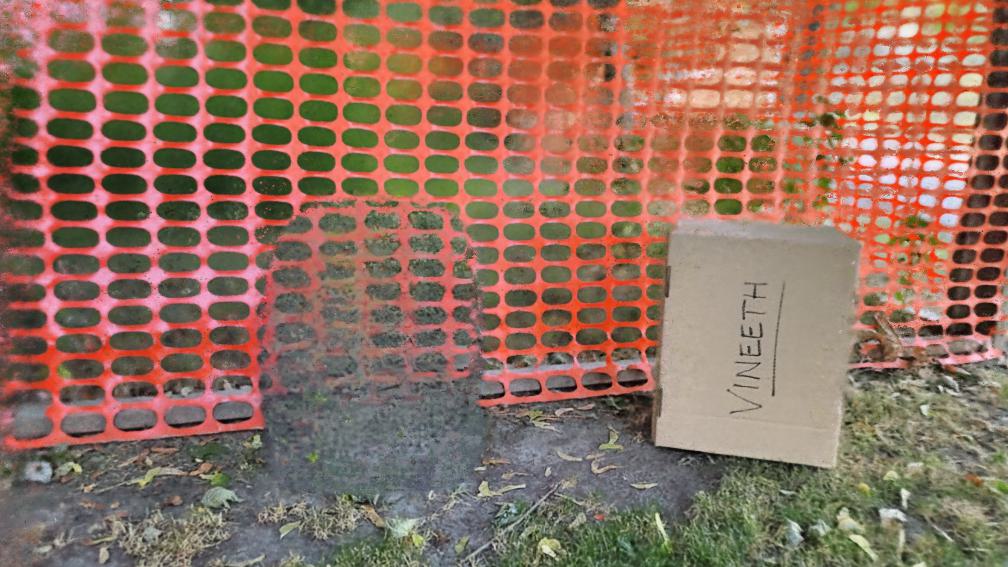} &
        \includegraphics[width=.24\linewidth,trim={0 0 7.11cm 4cm},clip]{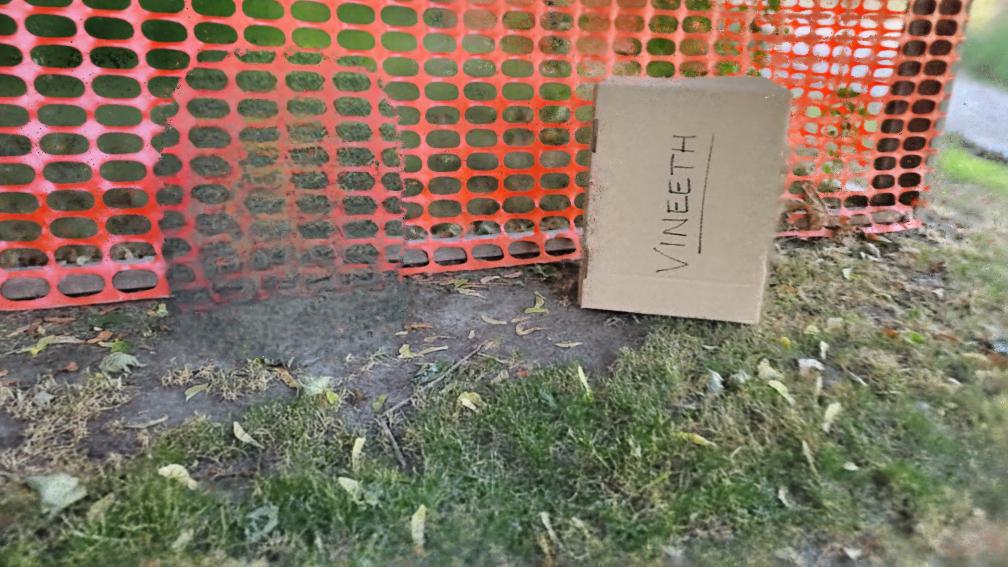} &
        \includegraphics[width=.24\linewidth,trim={0cm 0cm 0 0},clip]{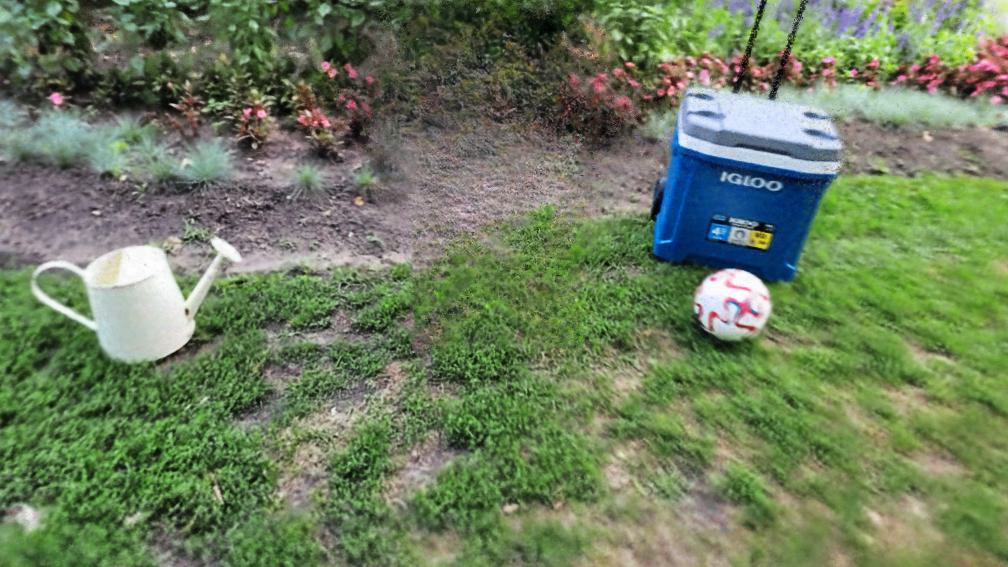} &
        \includegraphics[width=.24\linewidth,trim={0cm 0cm 0 0},clip]{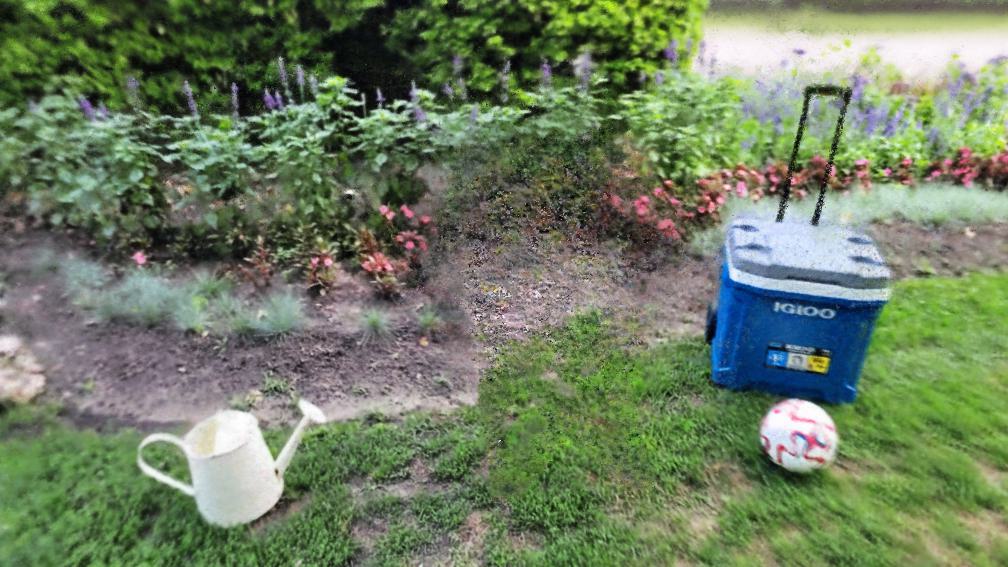} \\ [-0.25em]
        %
        \parbox[t]{1.2em}{\rotatebox[origin=l]{90}{\makecell{\scalebox{.55}{SPIn-NeRF (LDM)}}}} &
        \includegraphics[width=.24\linewidth,trim={0 0 7.11cm 4cm},clip]{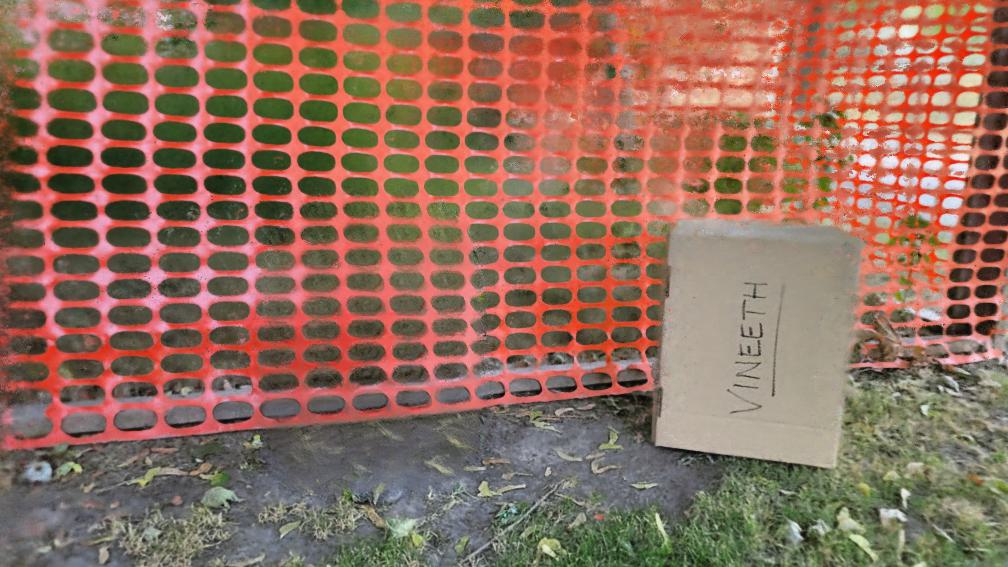} &
        \includegraphics[width=.24\linewidth,trim={0 0 7.11cm 4cm},clip]{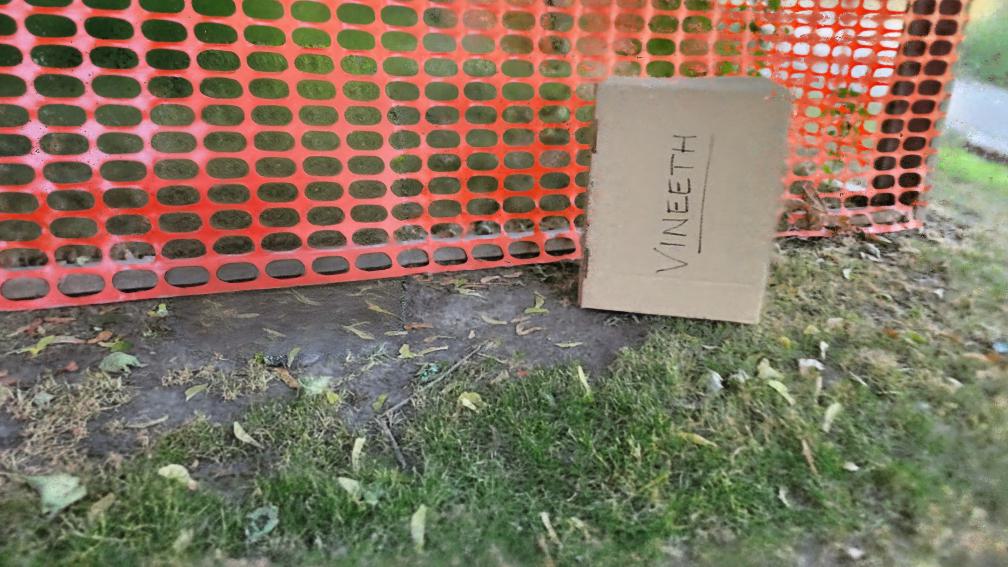} &
        \includegraphics[width=.24\linewidth,trim={0cm 0cm 0 0},clip]{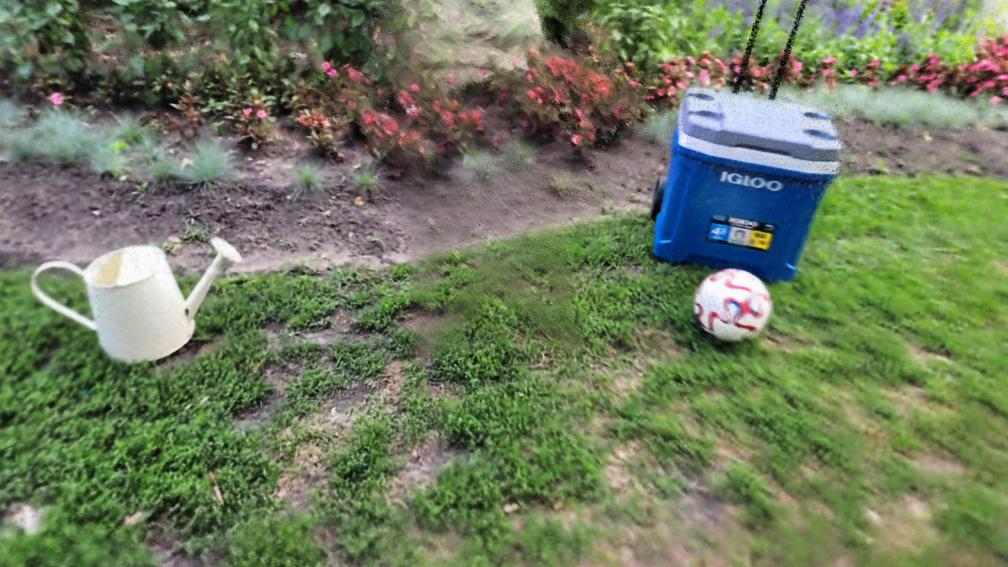} &
        \includegraphics[width=.24\linewidth,trim={0cm 0cm 0 0},clip]{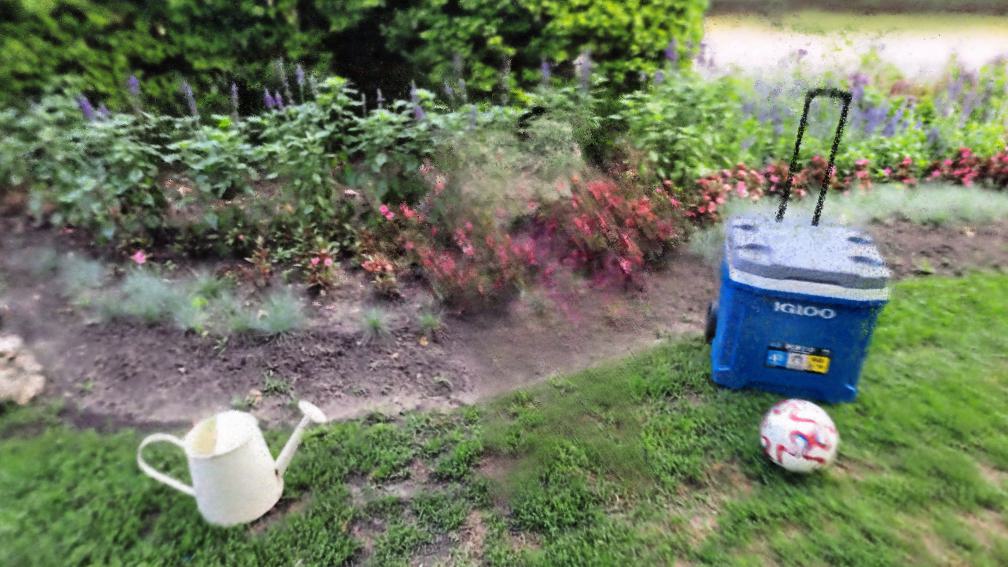} \\ [-0.25em]
        %
        \parbox[t]{1.2em}{\rotatebox[origin=l]{90}{\makecell{\scalebox{.55}{\hspace{.25em} InpaintNeRF360}}}} &
        \includegraphics[width=.24\linewidth,trim={0 0 7.11cm 4cm},clip]{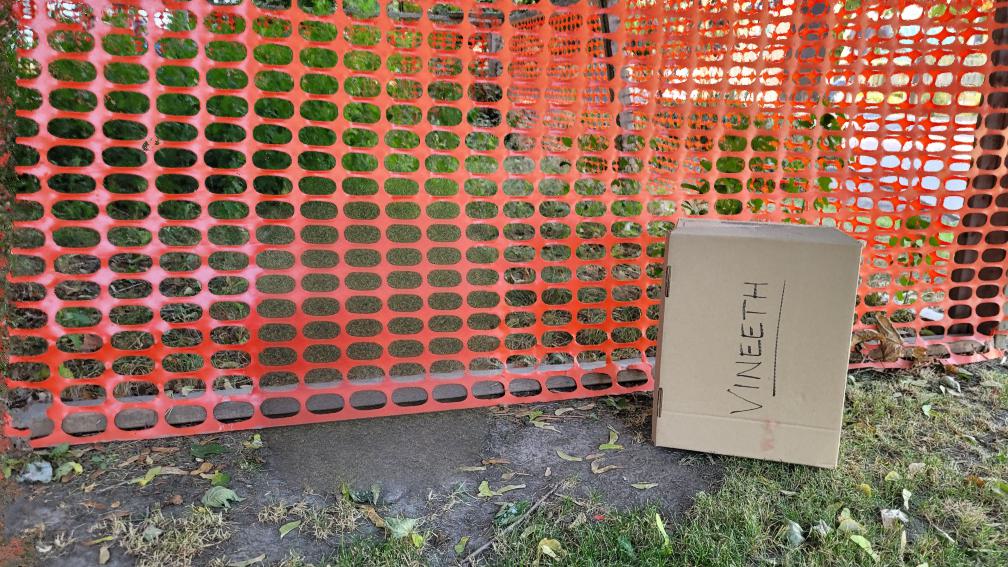} &
        \includegraphics[width=.24\linewidth,trim={0 0 7.11cm 4cm},clip]{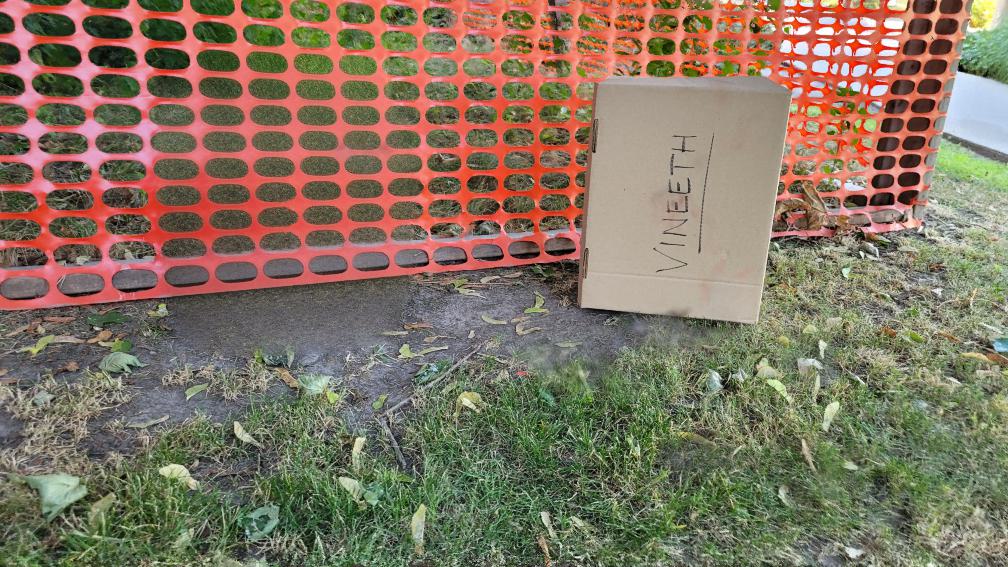} &
        \includegraphics[width=.24\linewidth,trim={0cm 0cm 0 0},clip]{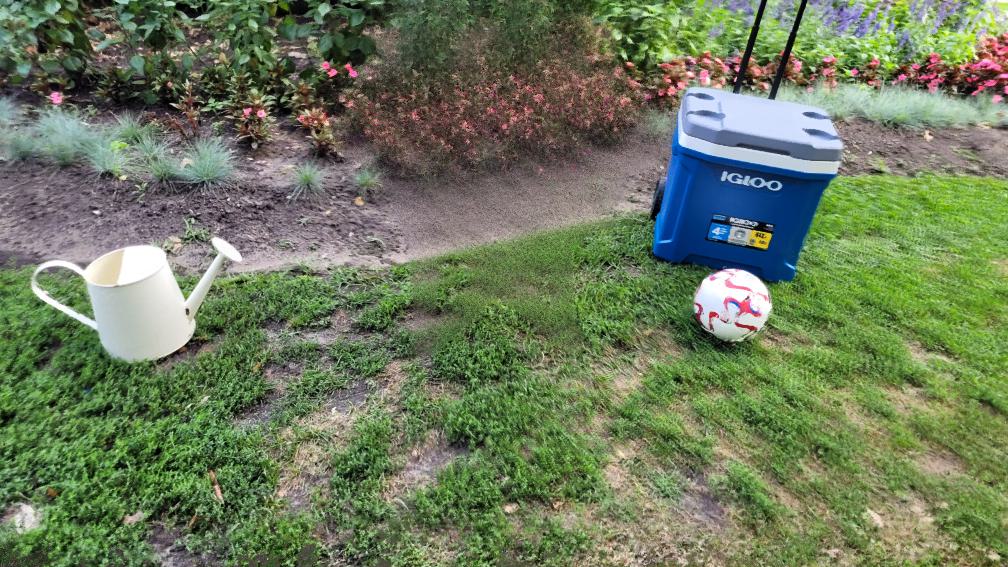} &
        \includegraphics[width=.24\linewidth,trim={0cm 0cm 0 0},clip]{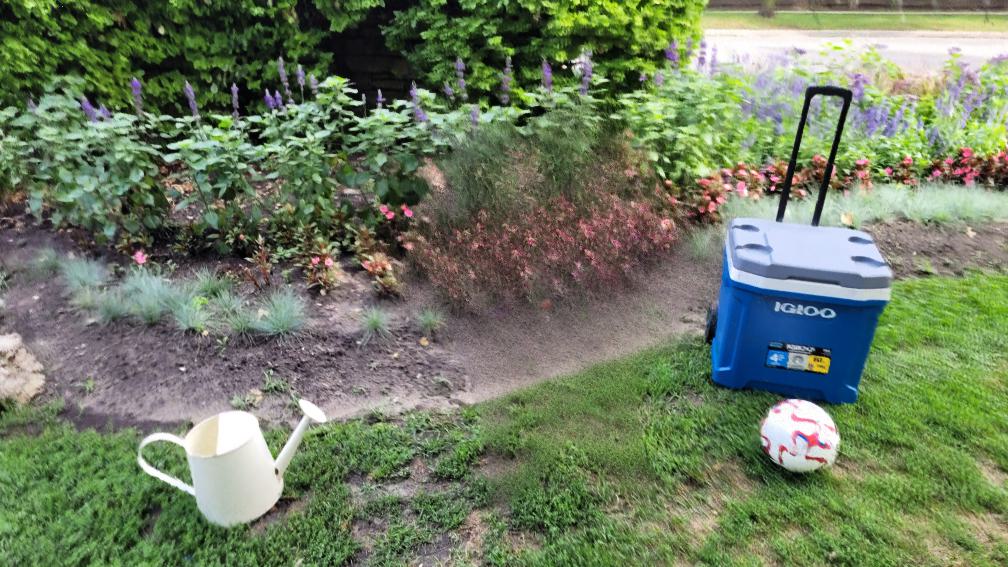}  \\[-0.25em]
        %
        \parbox[t]{1.2em}{\rotatebox[origin=l]{90}{\makecell{\scalebox{.55}{\hspace{1.5em} Inpaint3d}}}} &
        \includegraphics[width=.24\linewidth,trim={0 0 7.11cm 4cm},clip]{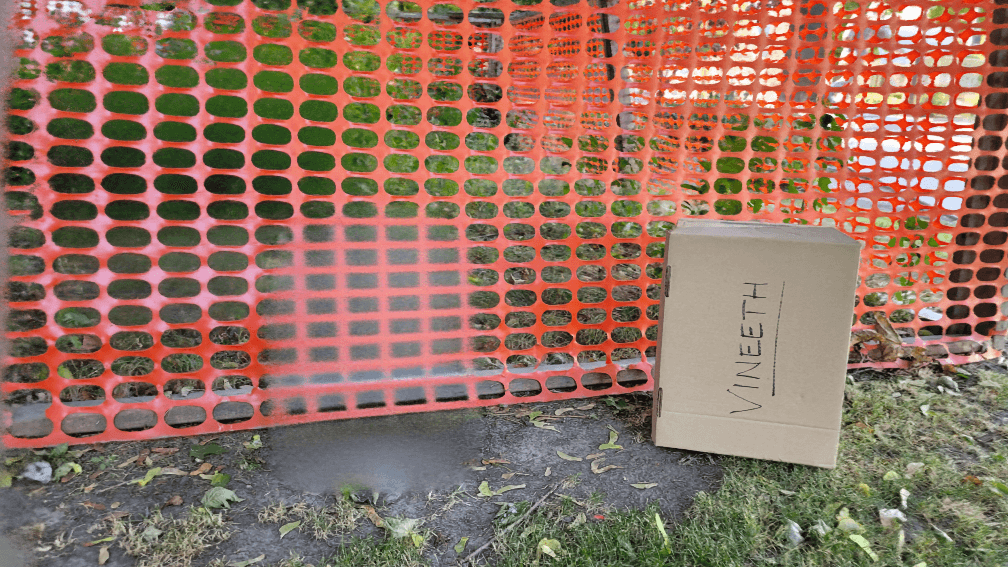} &
        \includegraphics[width=.24\linewidth,trim={0 0 7.11cm 4cm},clip]{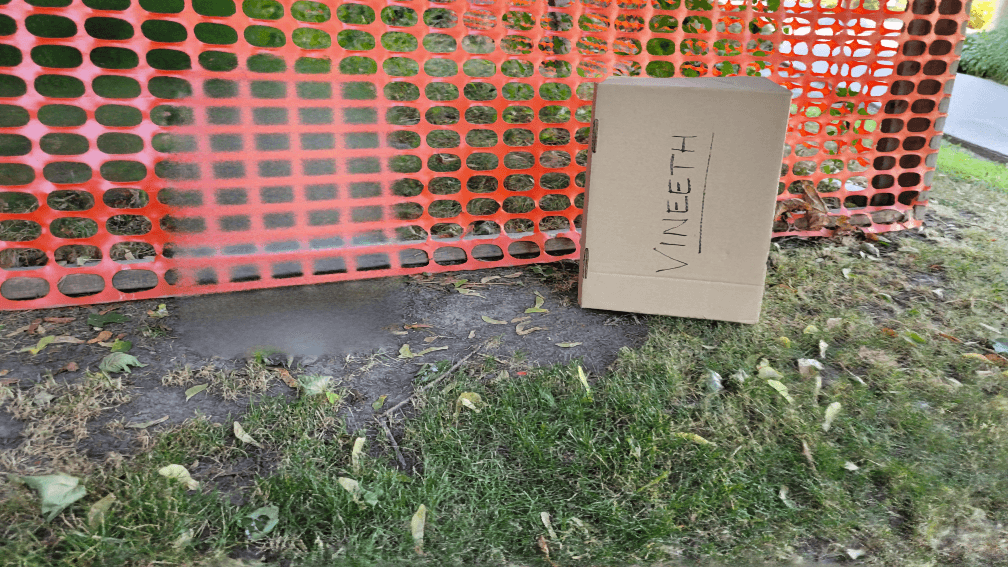} &
        \includegraphics[width=.24\linewidth,trim={0cm 0cm 0 0},clip]{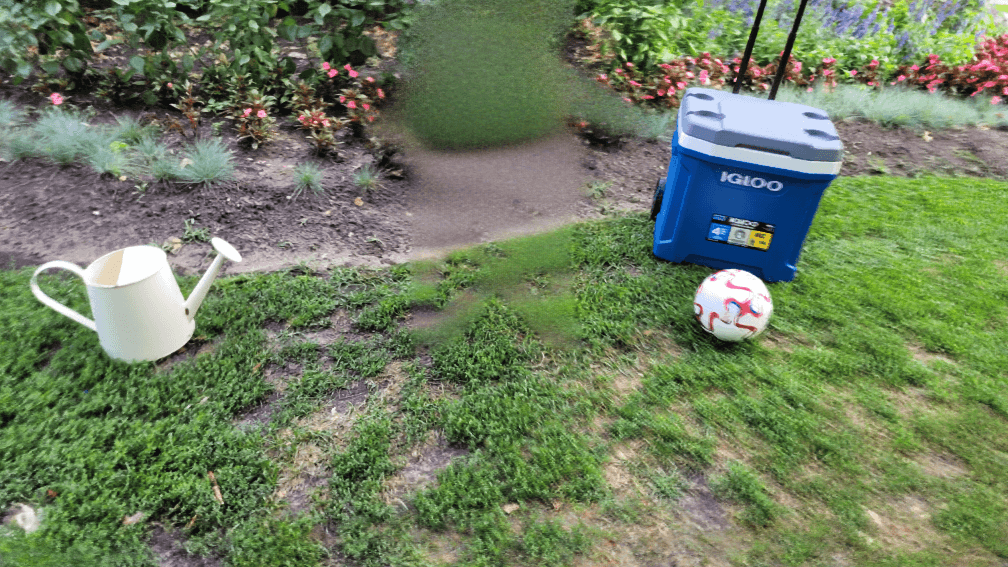} &
        \includegraphics[width=.24\linewidth,trim={0cm 0cm 0 0},clip]{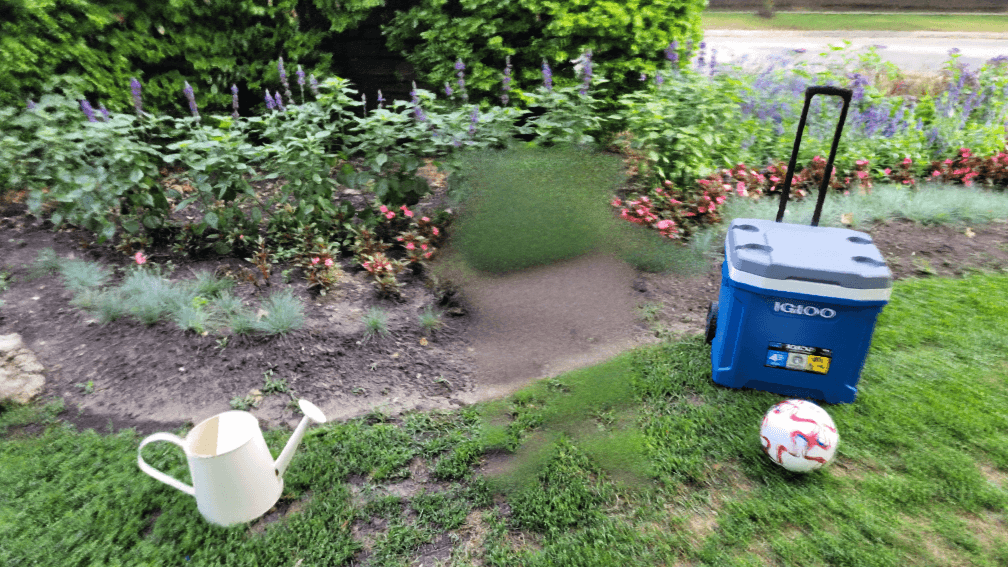} \\[-0.25em]
        %
        \parbox[t]{1.2em}{\rotatebox[origin=l]{90}{\makecell{\scalebox{.55}{\hspace{2.75em} Ours}}}} &
        \includegraphics[width=.24\linewidth,trim={0 0 7.11cm 4cm},clip]{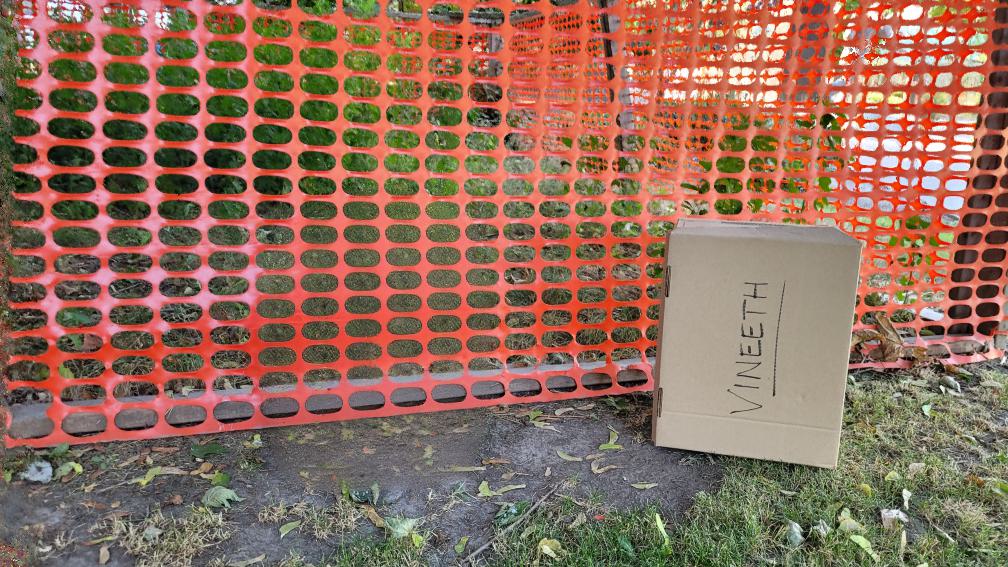} &
        \includegraphics[width=.24\linewidth,trim={0 0 7.11cm 4cm},clip]{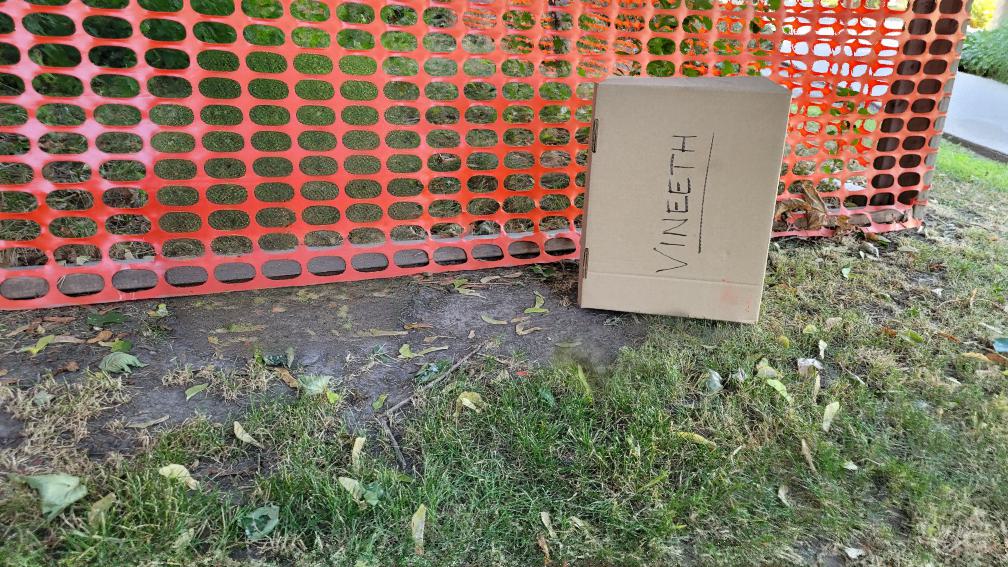} &
        \includegraphics[width=.24\linewidth,trim={0cm 0cm 0 0},clip]{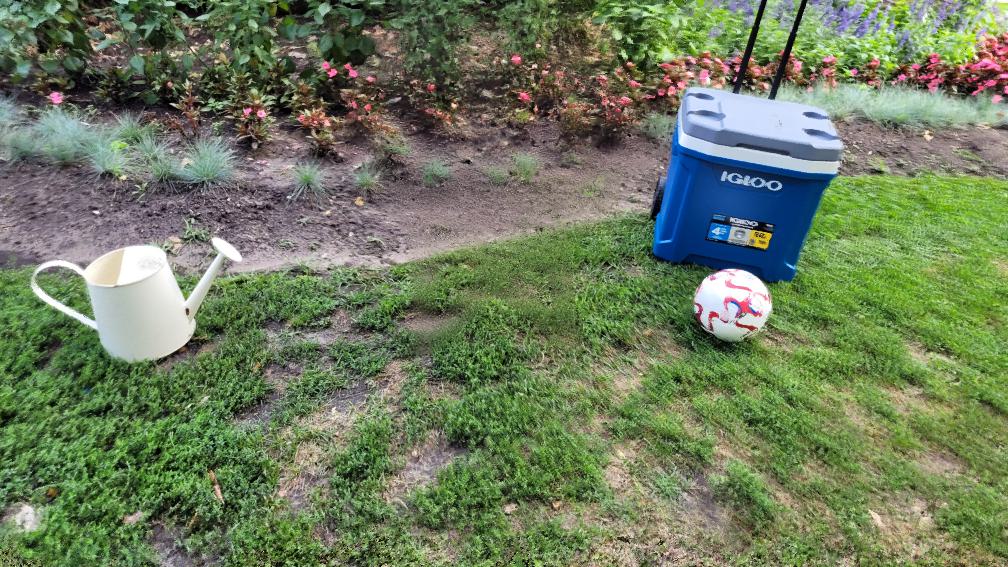} &
        \includegraphics[width=.24\linewidth,trim={0cm 0cm 0 0},clip]{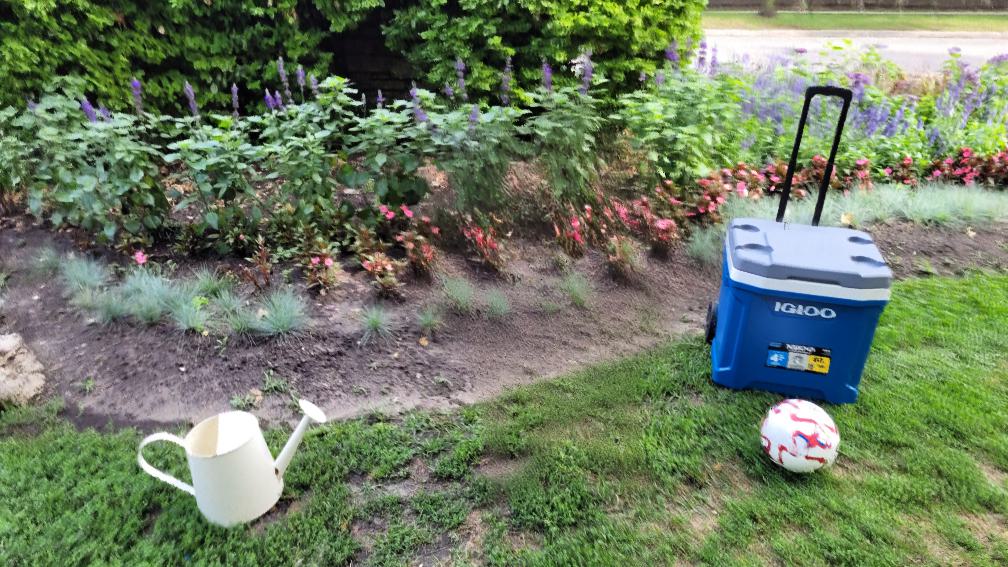}  \\[-0.25em]
        %
    \end{tabular}
    \vspace{-.5em}
    \caption{\textbf{Qualitative comparisons.} We present the results on the SPIn-NeRF~\cite{spinnerf} dataset.
    More results are shown in the supplementary material.
    }
    \label{fig:baselines}
    \vspace{-1em}
\end{figure}

\vspace{\subsecmargin}
\subsection{Per-Scene Finetuning}
\vspace{\subsecmargin}
In Figure~\ref{exp:ldm-finetune}, we qualitatively show the effectiveness of our per-scene finetuning on the latent diffusion model. 
Before the finetuning, our latent diffusion model inpaints arbitrary appearance, and even often time creates arbitrary objects in the inpainting region.
Such a high variation is a major issue causing the NeRF unable to converge, and often creates floaters and mist-like artifacts.
Note that we used a text prompt ``photo realistic, high quality, high resolution'' and a negative prompt~\cite{ho2021classifierfree} ``artifacts, low resolution, unknown, blur, low quality, human, animal, car.''
In contrast, after the finetuning, the inpainted results maintain a high consistency across individual inpainting results.
Note that we do not need extra prompts or negative prompts, since the finetuned model has learned scene-dependent tokens and finetuned the texture encoder with respect to the customized tokens.

\vspace{\subsecmargin}
\subsection{NeRF Inpainting}
\vspace{\subsecmargin}
\noindent \textbf{Qualitative analysis.} In Figure~\ref{fig:baselines}, we show the two most challenging scenes in the SPIn-NeRF dataset.
The scene with an orange net requires inpainting both the periodic textures of the net as well as the complex leaves on the ground.
Only our \modelName and InpaintNeRF360 are able to correctly complete the repetitive pattern of the net. 
Moreover, our method can further complete some of the leaves on the net, while other methods all converge to blurry appearances.
The second scene, removing the chair from a garden environment, involves completing complex plant textures at different granularity and patterns.
Only our \modelName can preserve the high-frequency details and seamlessly inpaint all plants by following the heterogeneous textures in the corresponding area.
All baseline methods converge to a blurry appearance due to the complexity of the texture and cross-view inconsistency from the inpainting prior.
In particular, it is worth noting that Inpaint3D trains their method on RealEstate10k~\cite{realestate}, which is one of the largest publicly available datasets for scene reconstruction.
The method shows outstanding performance in environments similar to the RealEstate10k distribution, but shows a significant performance drop in real-world environments, such as the SPIn-NeRF dataset.
This is potentially due to the limited dataset scale of RealEstate10k, and further reinforces the merits of utilizing a much more generalizable diffusion prior pretrained on the large-scale image dataset.
We include all visual comparisons in the supplementary material.

\noindent \textbf{Quantitative evaluation.} 
The quantitative analysis in Table~\ref{tab:quant} shows our method outperforms all methods in all metrics on both SPIn-NeRF and LLFF benchmarks.
Especially, the FID and KID metrics that focus on measuring the visual quality both indicate our method outperforms baselines by a large margin.
%

\begin{figure}[t!]
    \centering
    \setlength{\tabcolsep}{0pt}
    \begin{tabular}{c cc cc}
        \toprule
        \\ [-1.4em]
        \parbox[t]{1.6em}{\rotatebox[origin=l]{90}{\scalebox{.5}{\hspace{1.5em}\makecell{ Input$+$ Mask}}}} &
        \includegraphics[width=.24\linewidth,trim={8.7cm 6cm 9cm 4cm},clip]{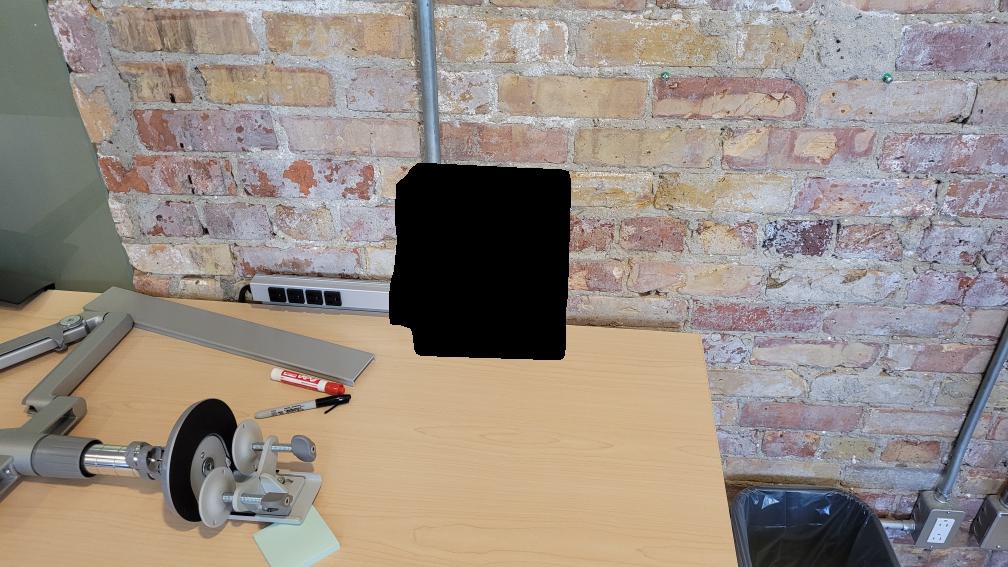} &
        \includegraphics[width=.24\linewidth,trim={8.7cm 6cm 9cm 4cm},clip]{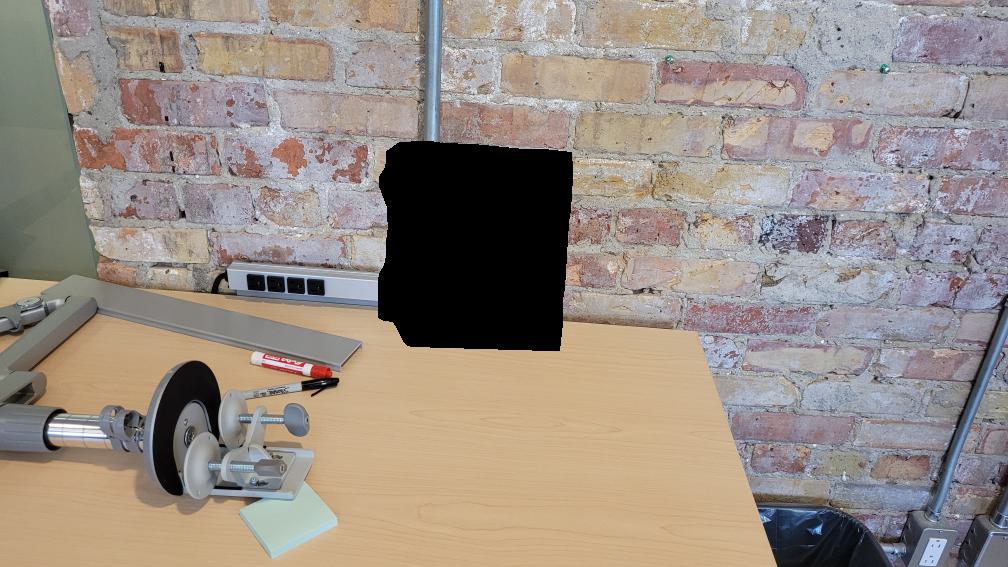} &
        \includegraphics[width=.24\linewidth,trim={5.5cm 8cm 8.7cm 0cm},clip]{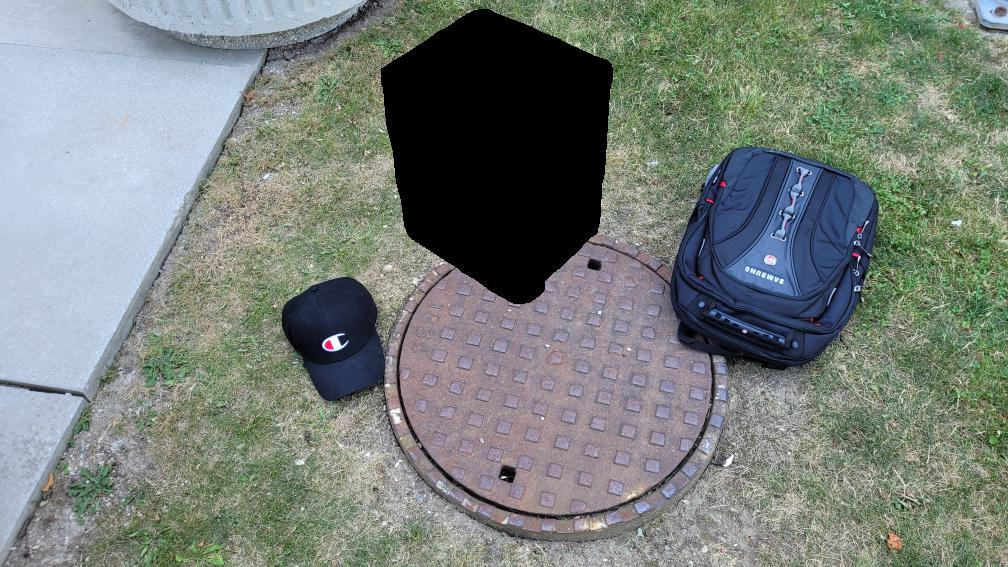} &
        \includegraphics[width=.24\linewidth,trim={7.5cm 8cm 6.7cm 0cm},clip]{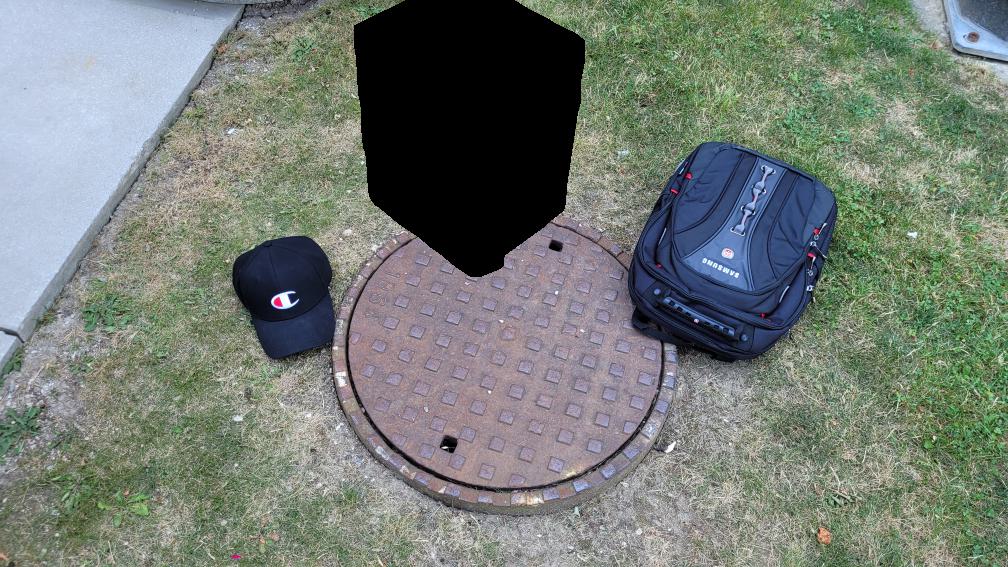} \\ [-0.3em]
        \midrule
        \parbox[t]{1.6em}{\rotatebox[origin=l]{90}{\scalebox{.5}{\hspace{2.25em}\makecell{Ours $-$Adv \\ $+$ I-Recon}}}} &
        \includegraphics[width=.24\linewidth,trim={8.7cm 6cm 9cm 4cm},clip]{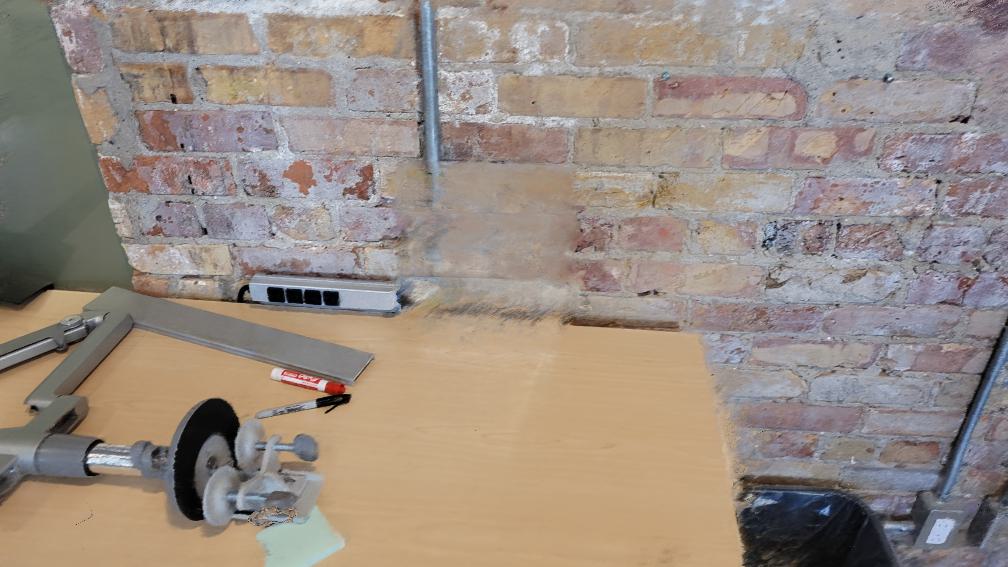} &
        \includegraphics[width=.24\linewidth,trim={8.7cm 6cm 9cm 4cm},clip]{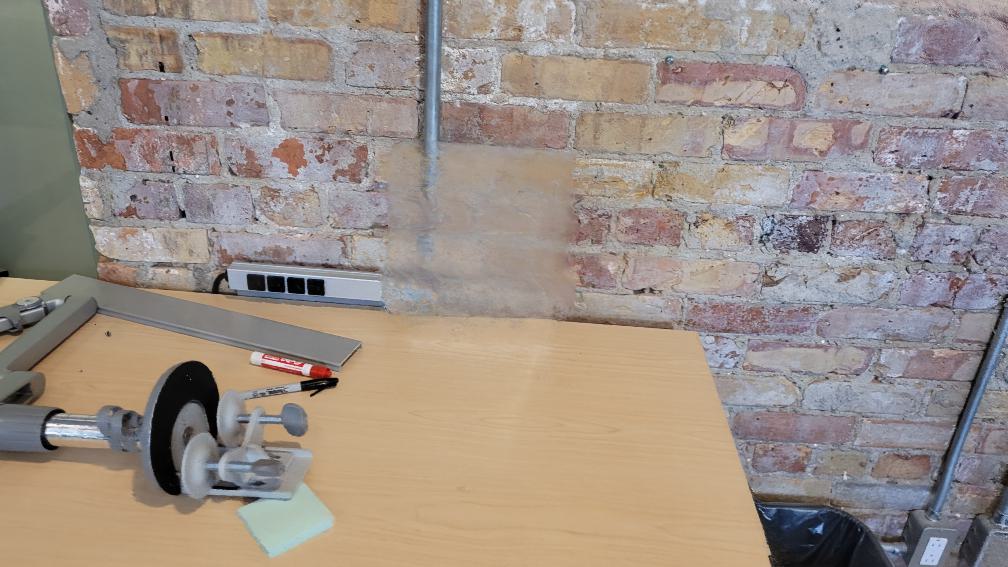} &
        \includegraphics[width=.24\linewidth,trim={5.5cm 8cm 8.7cm 0cm},clip]{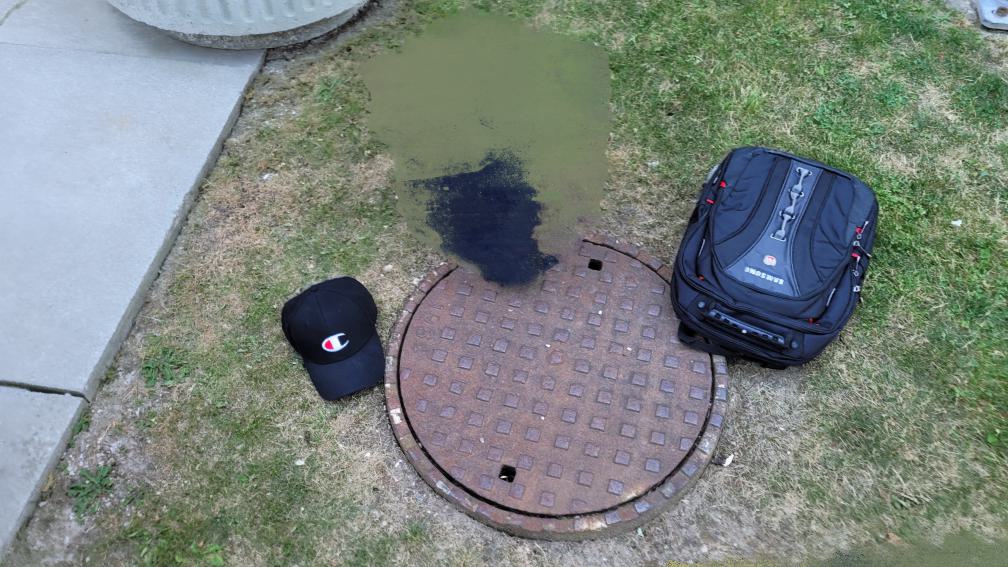} &
        \includegraphics[width=.24\linewidth,trim={7.5cm 8cm 6.7cm 0cm},clip]{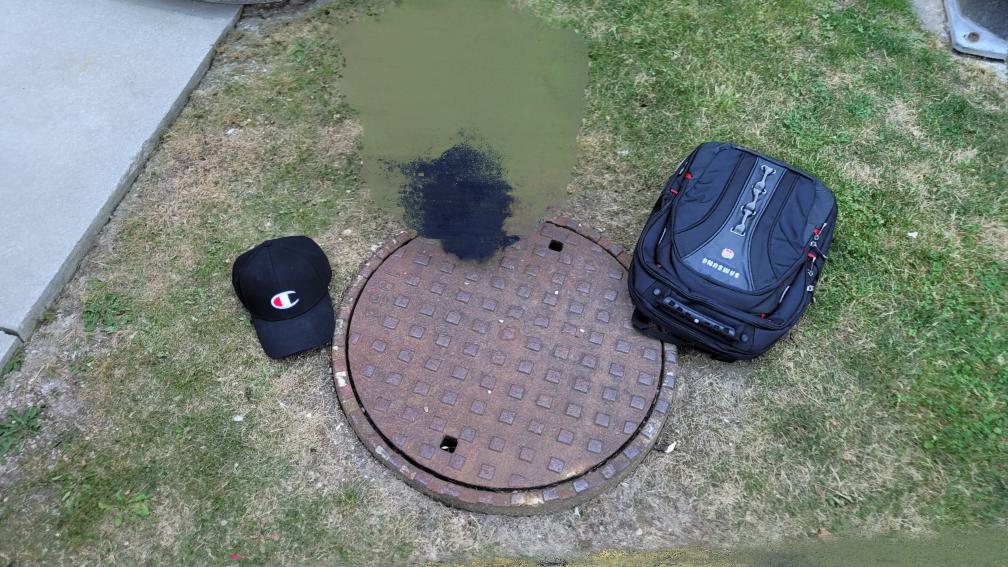} \\ [-0.3em]
        %
        \parbox[t]{1.6em}{\rotatebox[origin=l]{90}{\scalebox{.5}{\hspace{2.5em}\makecell{Ours $-$Adv \\ $+$ LPIPS}}}} &
        \includegraphics[width=.24\linewidth,trim={8.7cm 6cm 9cm 4cm},clip]{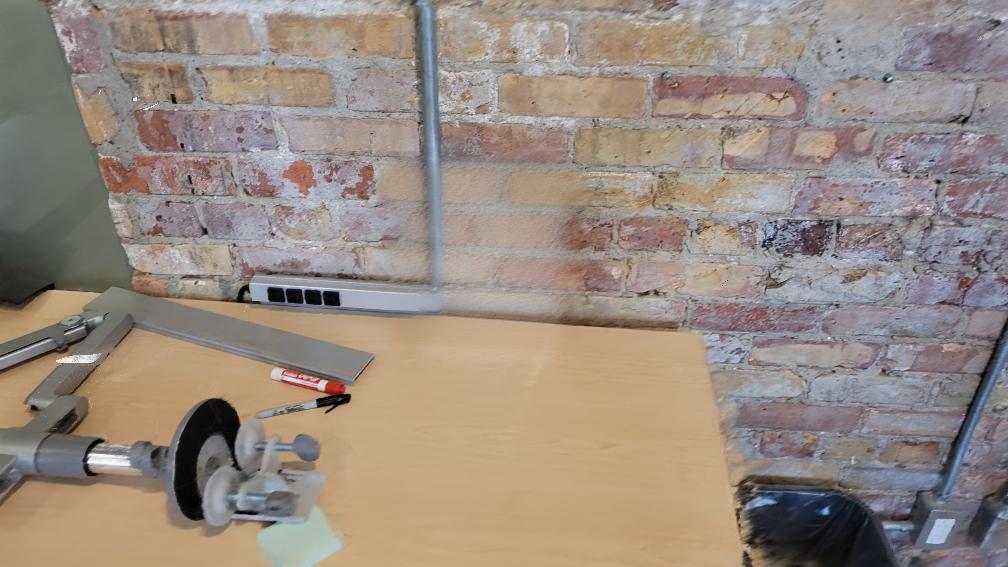} &
        \includegraphics[width=.24\linewidth,trim={8.7cm 6cm 9cm 4cm},clip]{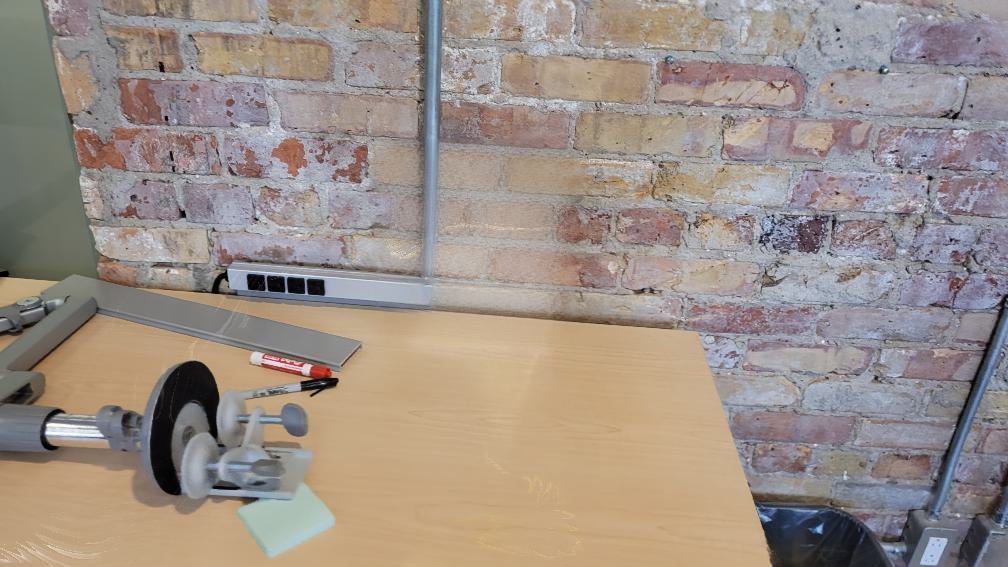} &
        \includegraphics[width=.24\linewidth,trim={5.5cm 8cm 8.7cm 0cm},clip]{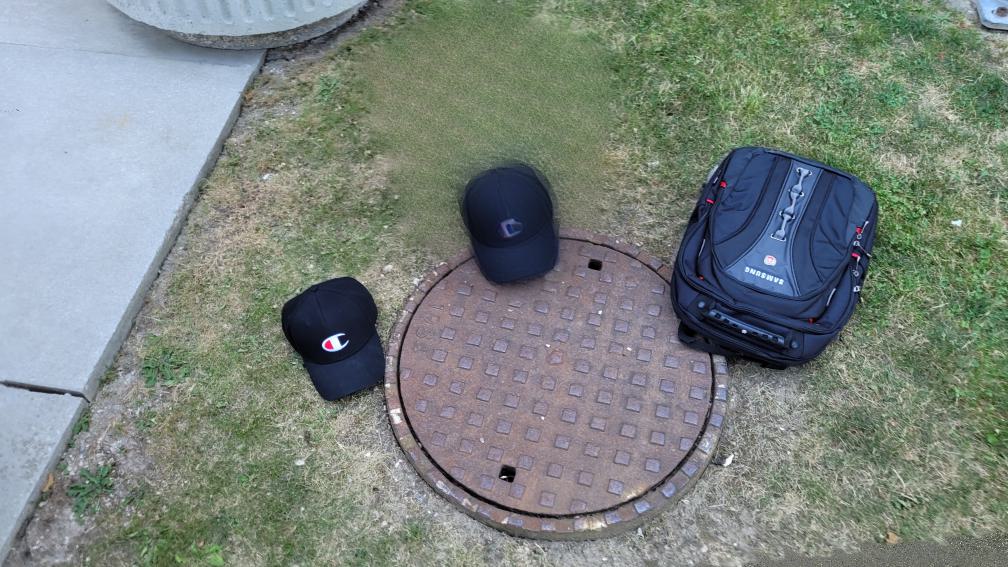} &
        \includegraphics[width=.24\linewidth,trim={7.5cm 8cm 6.7cm 0cm},clip]{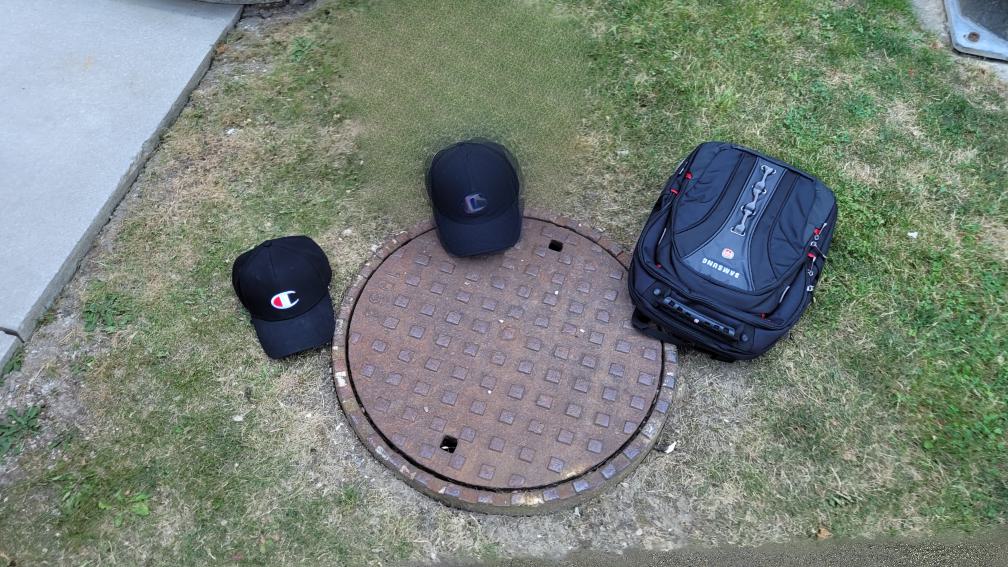} \\ [-0.3em]
        %
        \parbox[t]{1.6em}{\rotatebox[origin=l]{90}{\scalebox{.5}{\hspace{.25em}\makecell{ Ours $-$Adv \\ $+$I-Recon$+$ LPIPS }}}} &
        \includegraphics[width=.24\linewidth,trim={8.7cm 6cm 9cm 4cm},clip]{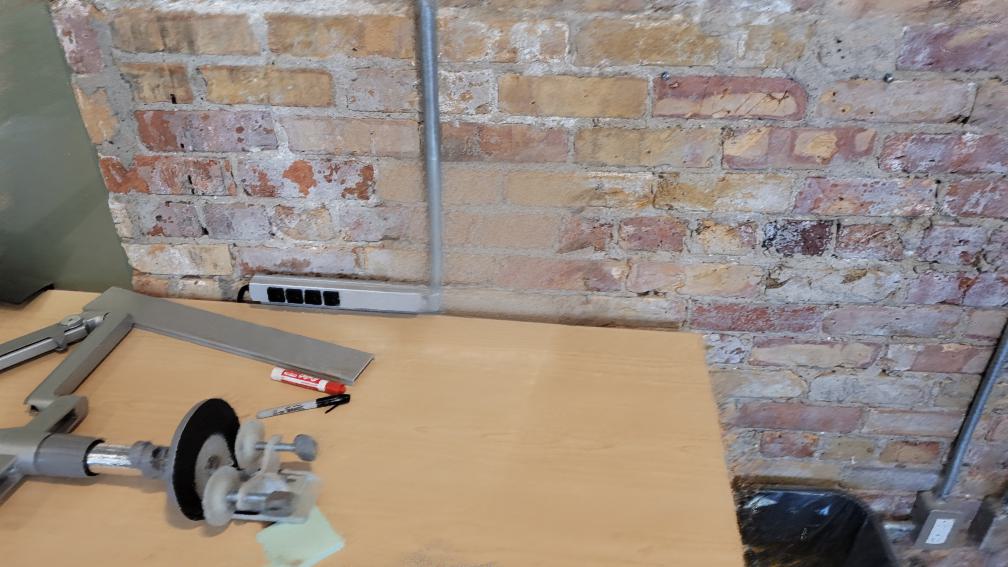} &
        \includegraphics[width=.24\linewidth,trim={8.7cm 6cm 9cm 4cm},clip]{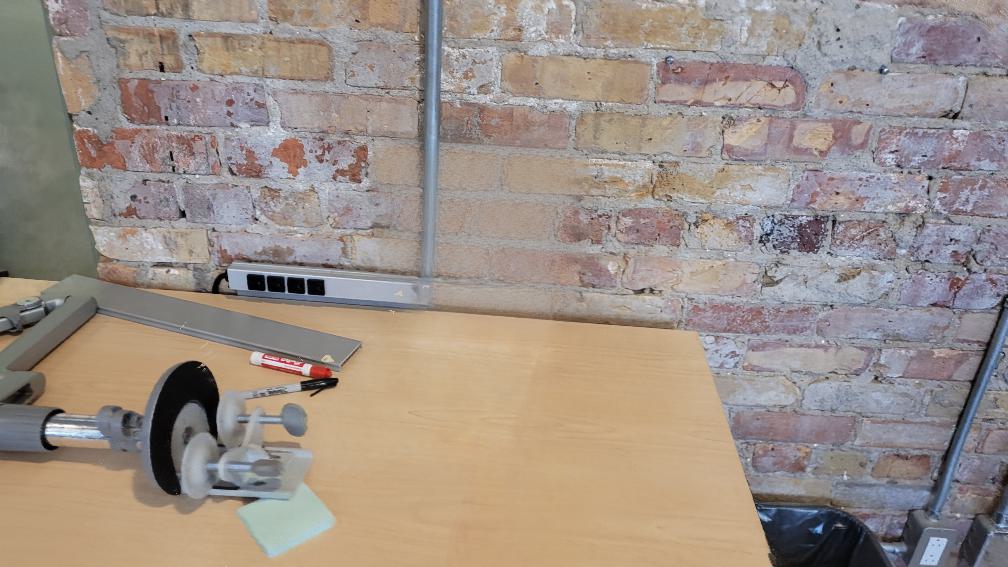} &
        \includegraphics[width=.24\linewidth,trim={5.5cm 8cm 8.7cm 0cm},clip]{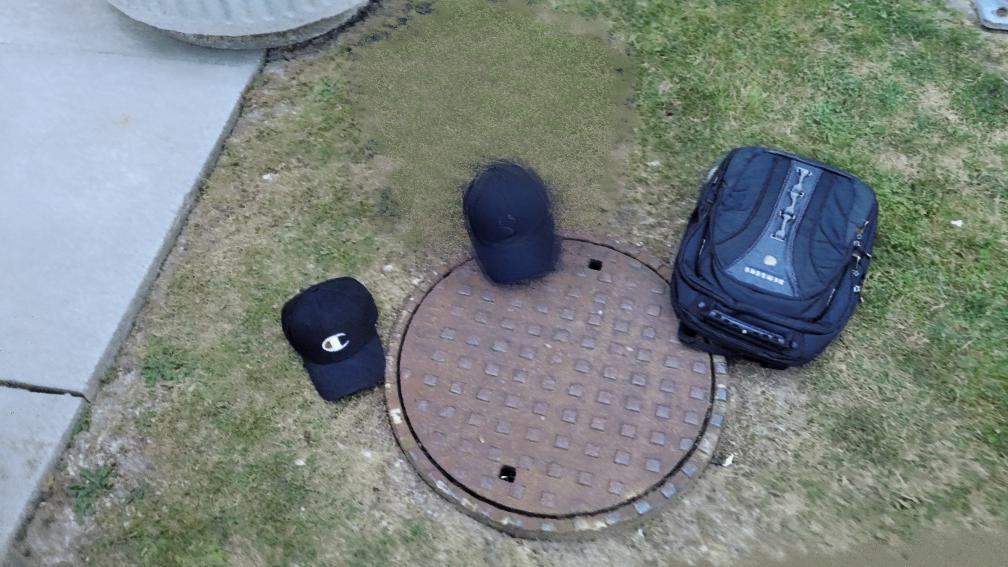} &
        \includegraphics[width=.24\linewidth,trim={7.5cm 8cm 6.7cm 0cm},clip]{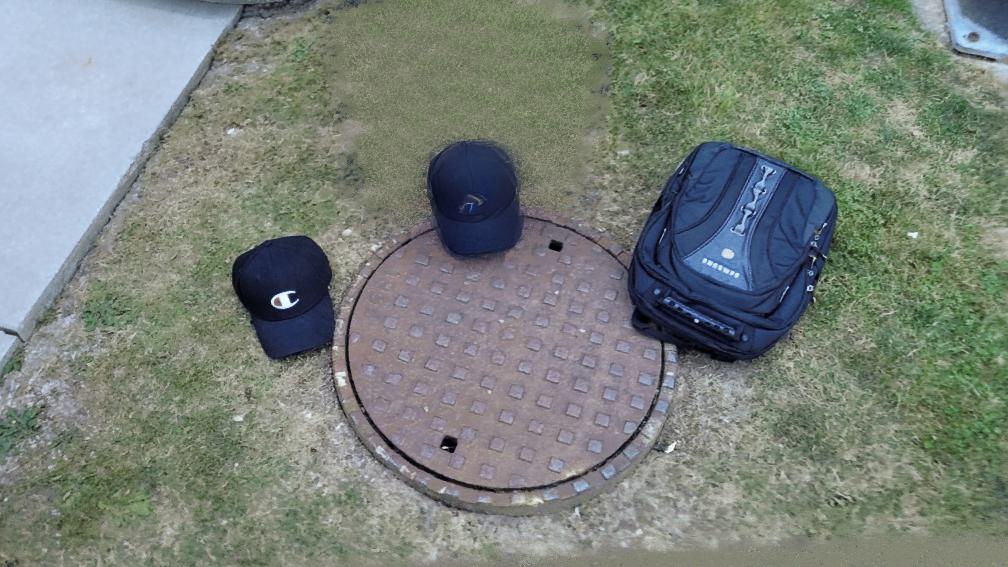} \\ [-0.3em]
        \midrule
        \parbox[t]{1.6em}{\rotatebox[origin=l]{90}{\scalebox{.5}{\hspace{2.5em}\makecell{ Ours \\$+$ I-Recon  }}}} &
        \includegraphics[width=.24\linewidth,trim={8.7cm 6cm 9cm 4cm},clip]{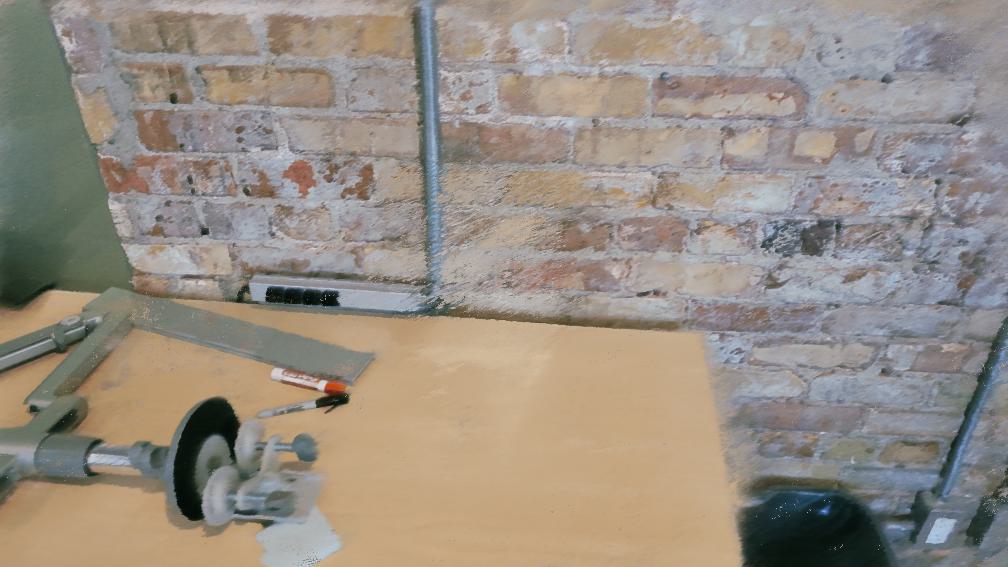} &
        \includegraphics[width=.24\linewidth,trim={8.7cm 6cm 9cm 4cm},clip]{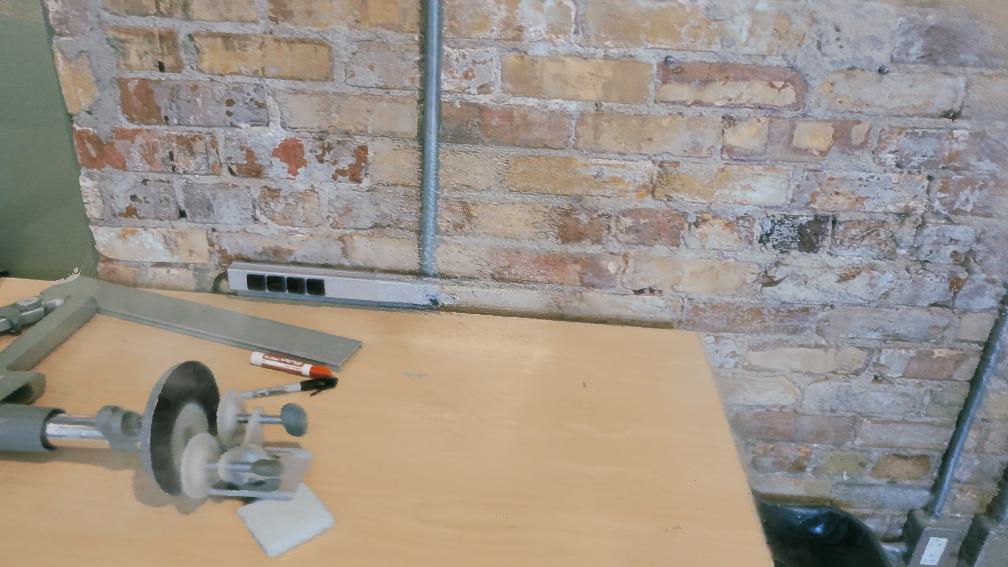} &
        \includegraphics[width=.24\linewidth,trim={5.5cm 8cm 8.7cm 0cm},clip]{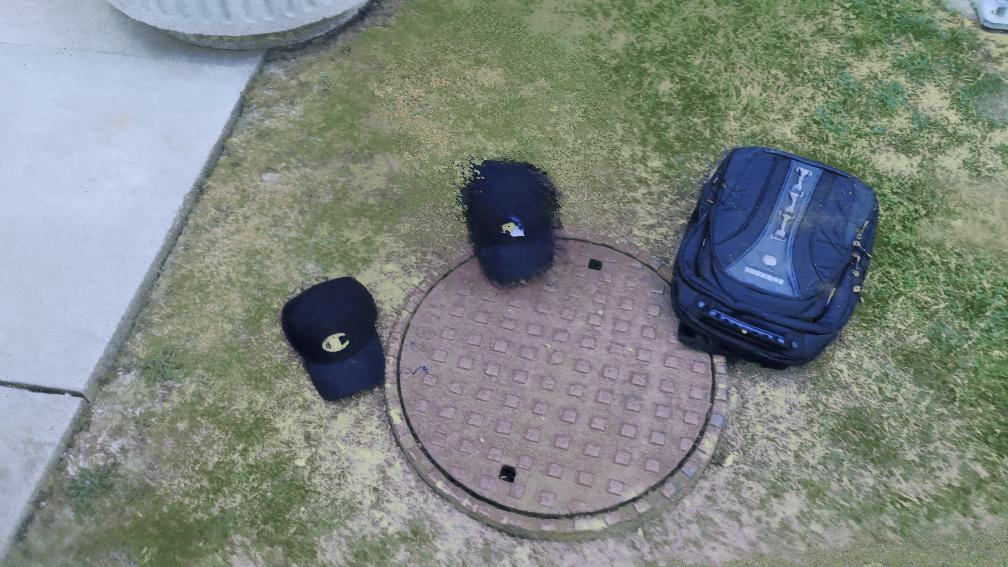} &
        \includegraphics[width=.24\linewidth,trim={7.5cm 8cm 6.7cm 0cm},clip]{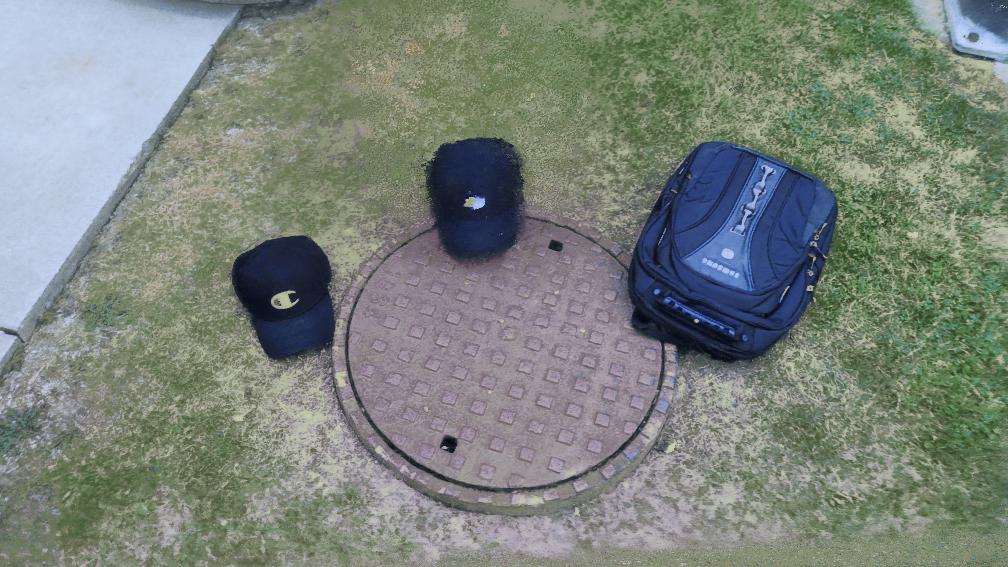} \\ [-0.3em]
        %
        \parbox[t]{1.6em}{\rotatebox[origin=l]{90}{\scalebox{.5}{\hspace{3em}\makecell{ Ours \\$+$ LPIPS}}}} &
        \includegraphics[width=.24\linewidth,trim={8.7cm 6cm 9cm 4cm},clip]{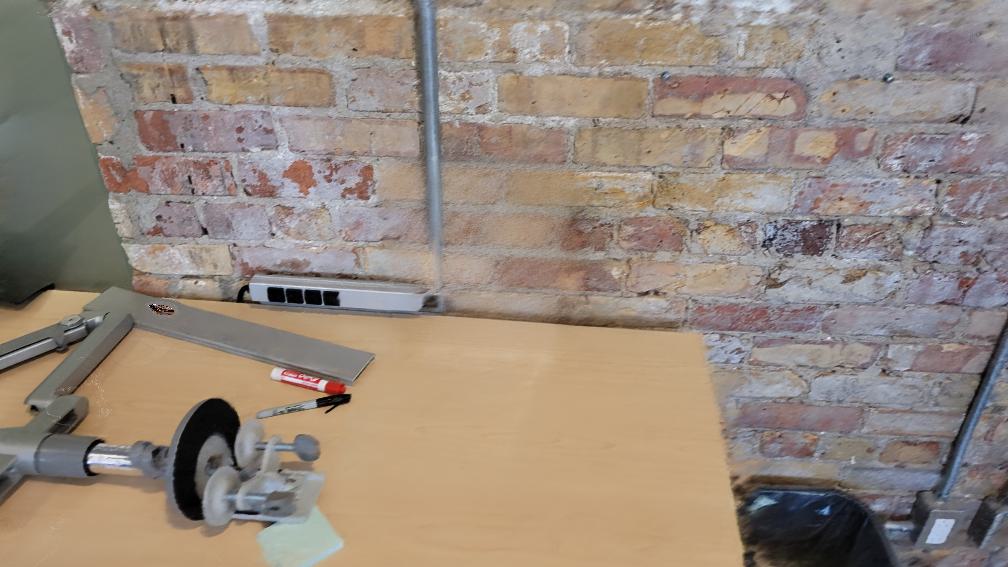} &
        \includegraphics[width=.24\linewidth,trim={8.7cm 6cm 9cm 4cm},clip]{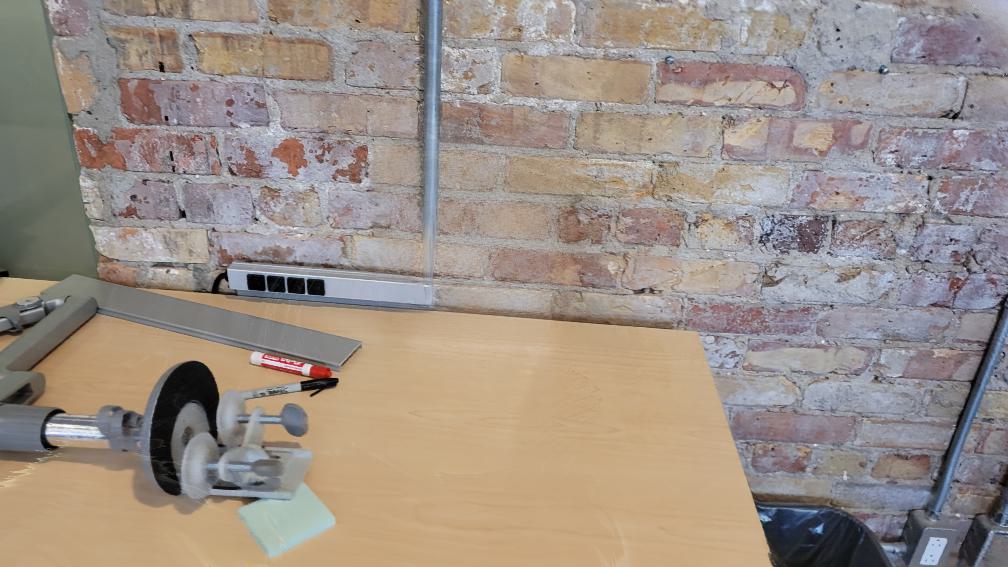} &
        \includegraphics[width=.24\linewidth,trim={5.5cm 8cm 8.7cm 0cm},clip]{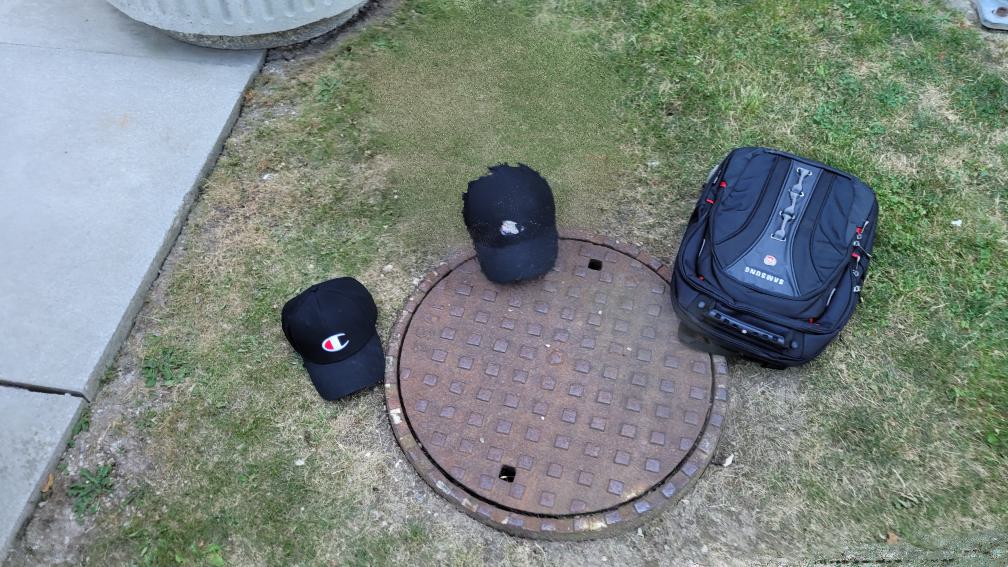} &
        \includegraphics[width=.24\linewidth,trim={7.5cm 8cm 6.7cm 0cm},clip]{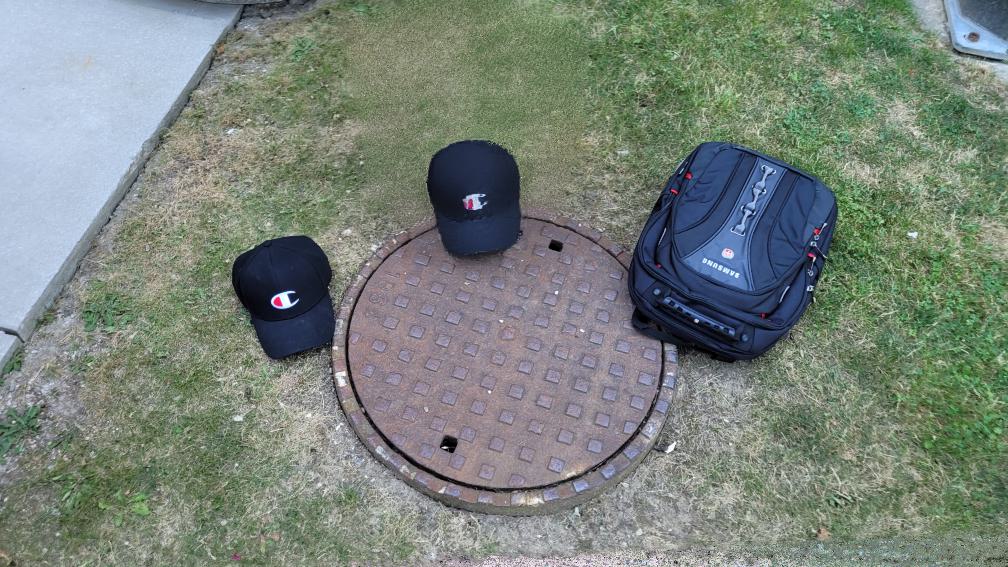} \\ [-0.3em]
        %
        \parbox[t]{1.6em}{\rotatebox[origin=l]{90}{\scalebox{.5}{\hspace{1em}\makecell{ Ours $-$Per-Scene \\ Customization}}}} &
        \includegraphics[width=.24\linewidth,trim={8.7cm 6cm 9cm 4cm},clip]{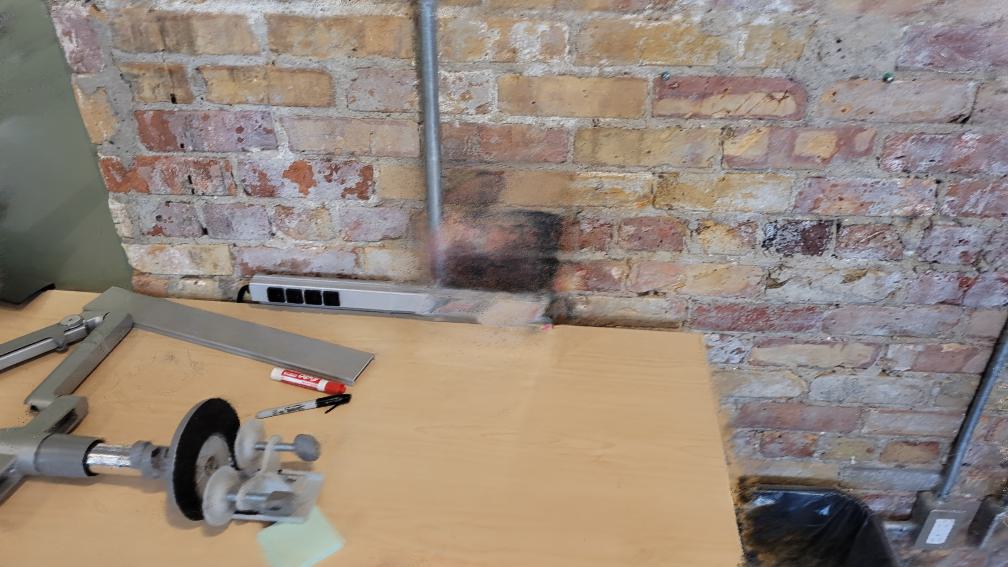} &
        \includegraphics[width=.24\linewidth,trim={8.7cm 6cm 9cm 4cm},clip]{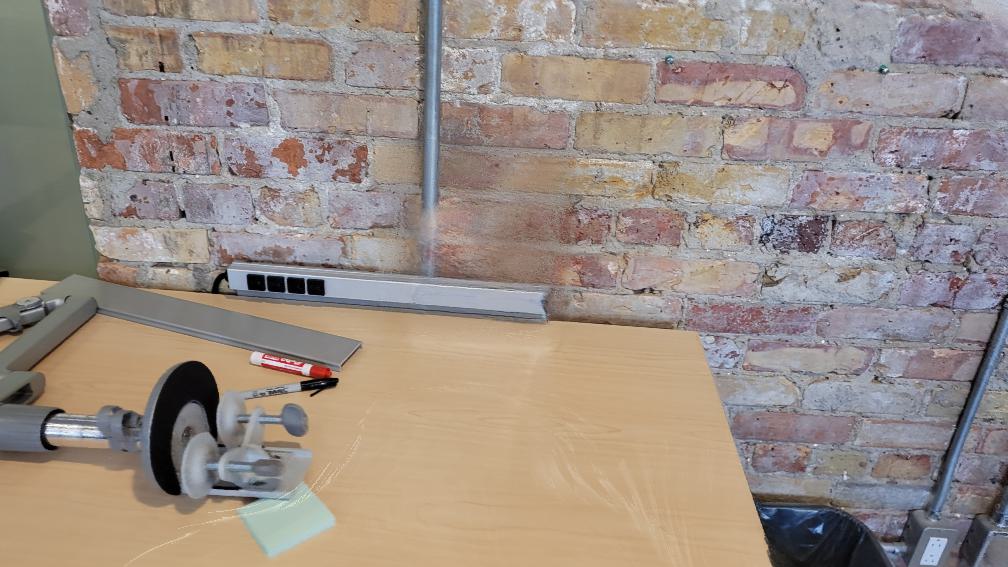} &
        \includegraphics[width=.24\linewidth,trim={5.5cm 8cm 8.7cm 0cm},clip]{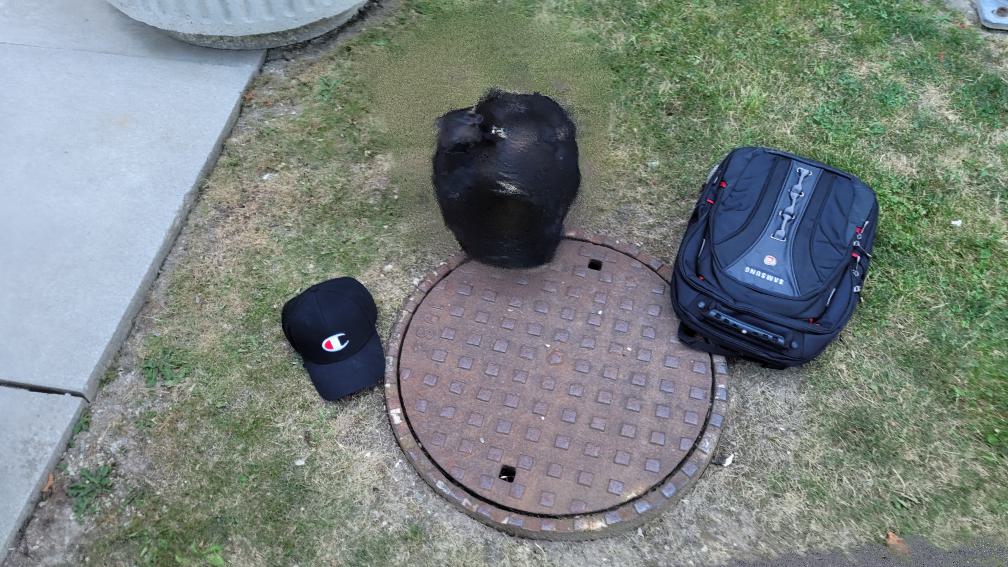} &
        \includegraphics[width=.24\linewidth,trim={7.5cm 8cm 6.7cm 0cm},clip]{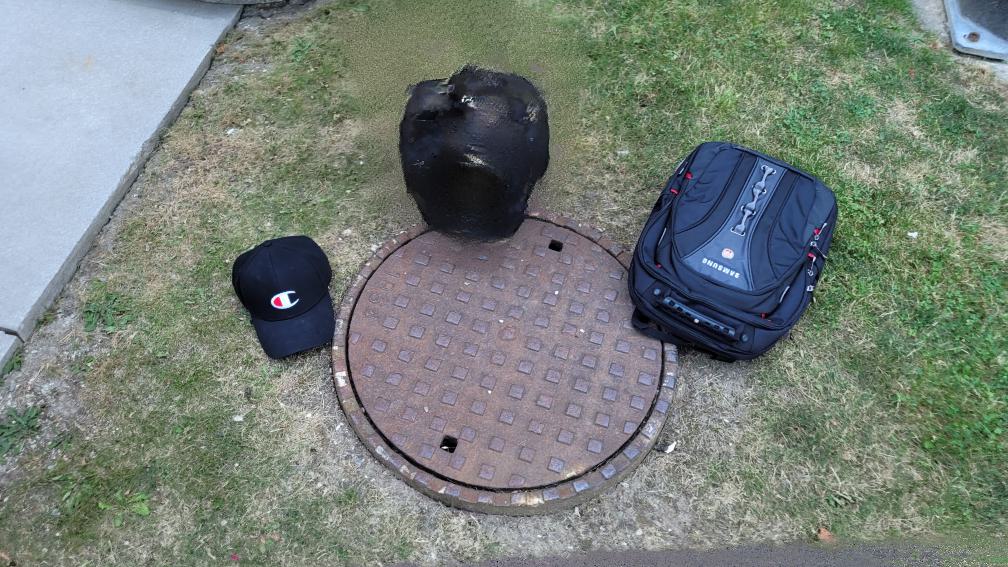} \\ [-0.3em]
        %
        \parbox[t]{1.6em}{\rotatebox[origin=l]{90}{\scalebox{.5}{\hspace{.25em} \makecell{Ours \\ $-$Feature Matching}}}} &
        \includegraphics[width=.24\linewidth,trim={8.7cm 6cm 9cm 4cm},clip]{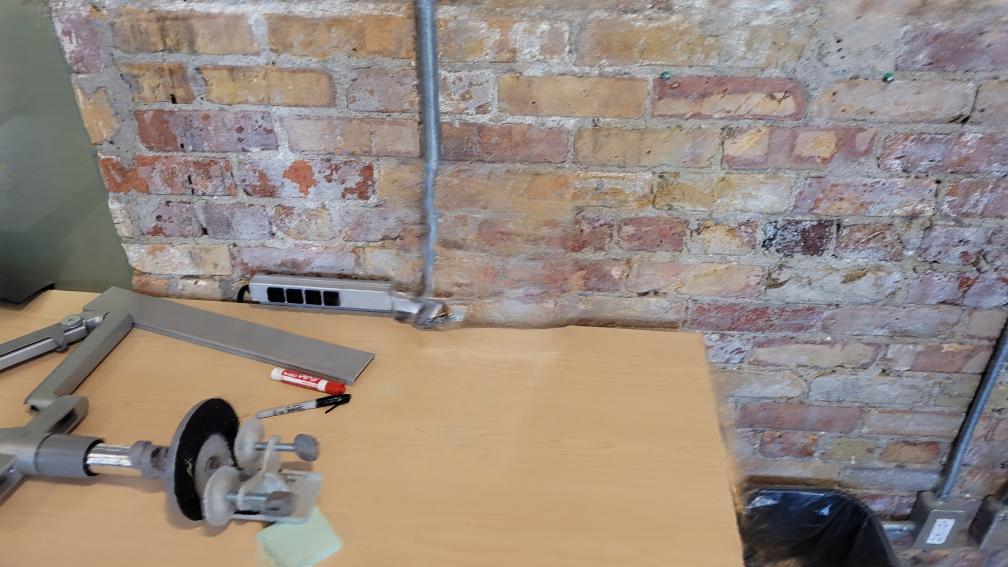} &
        \includegraphics[width=.24\linewidth,trim={8.7cm 6cm 9cm 4cm},clip]{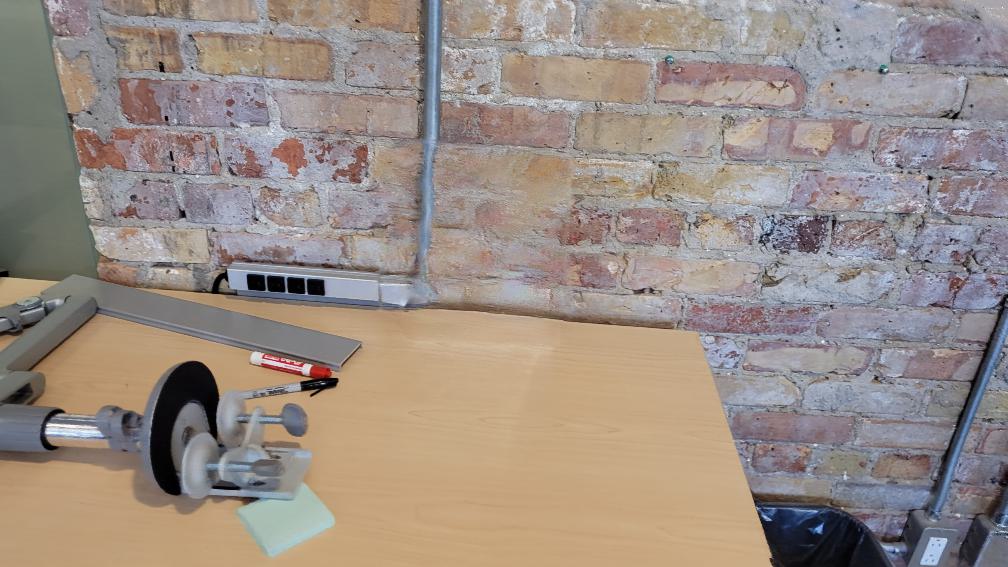} &
        \includegraphics[width=.24\linewidth,trim={5.5cm 8cm 8.7cm 0cm},clip]{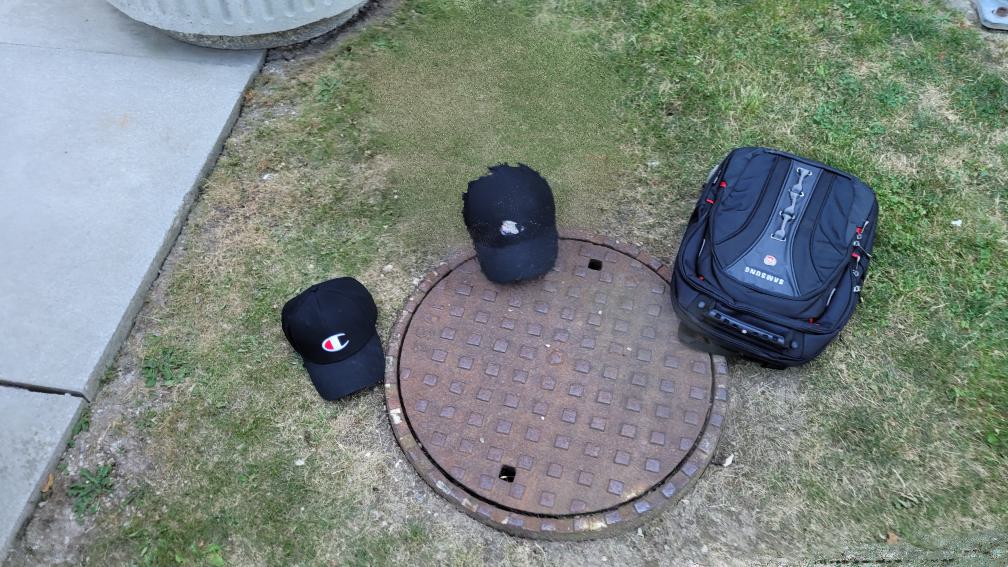} &
        \includegraphics[width=.24\linewidth,trim={7.5cm 8cm 6.7cm 0cm},clip]{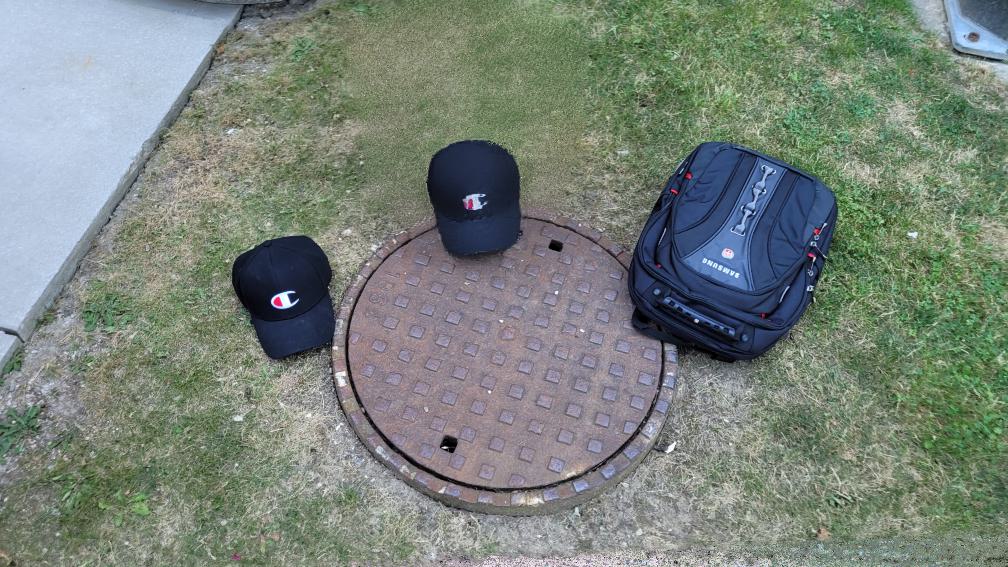} \\ [-0.3em]
        %
        \parbox[t]{1.6em}{\rotatebox[origin=l]{90}{\scalebox{.5}{\hspace{1em} \makecell{Ours \\ $-$Adv Masking}}}} &
        \includegraphics[width=.24\linewidth,trim={8.7cm 6cm 9cm 4cm},clip]{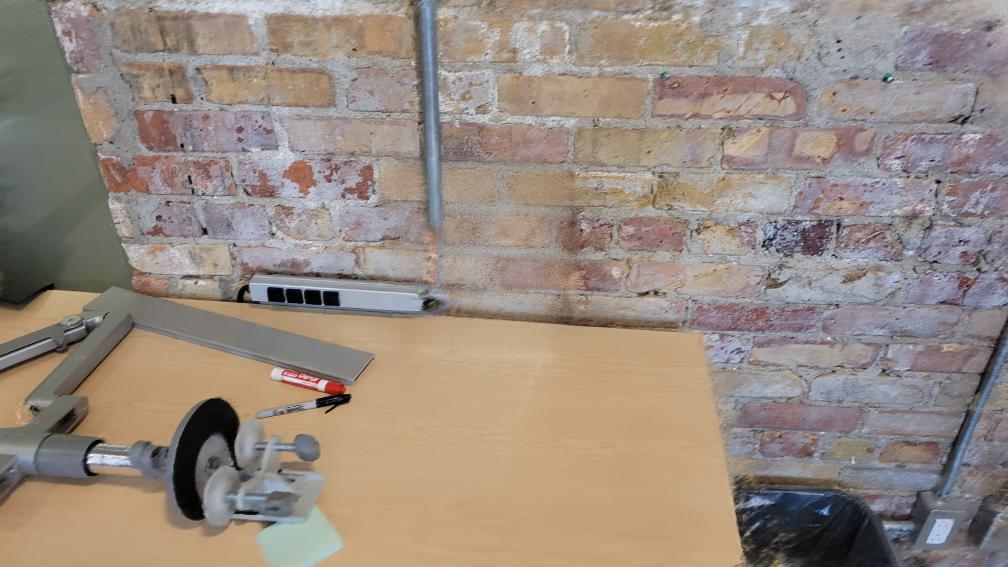} &
        \includegraphics[width=.24\linewidth,trim={8.7cm 6cm 9cm 4cm},clip]{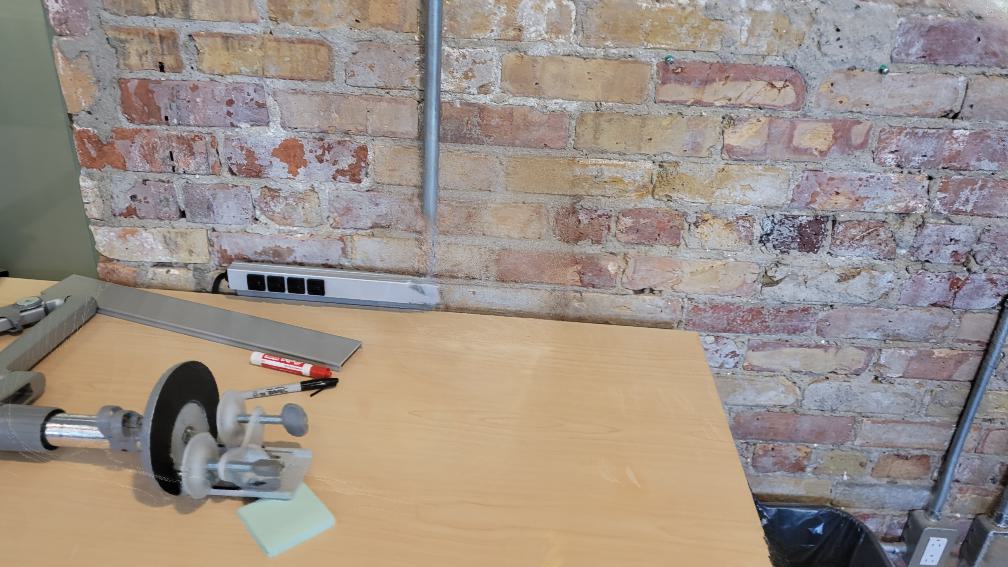} &
        \includegraphics[width=.24\linewidth,trim={5.5cm 8cm 8.7cm 0cm},clip]{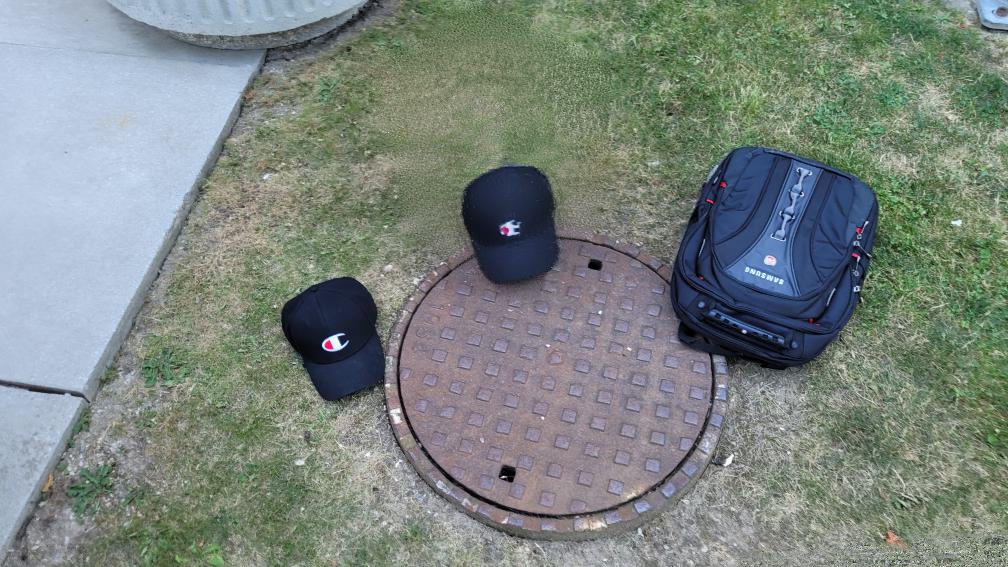} &
        \includegraphics[width=.24\linewidth,trim={7.5cm 8cm 6.7cm 0cm},clip]{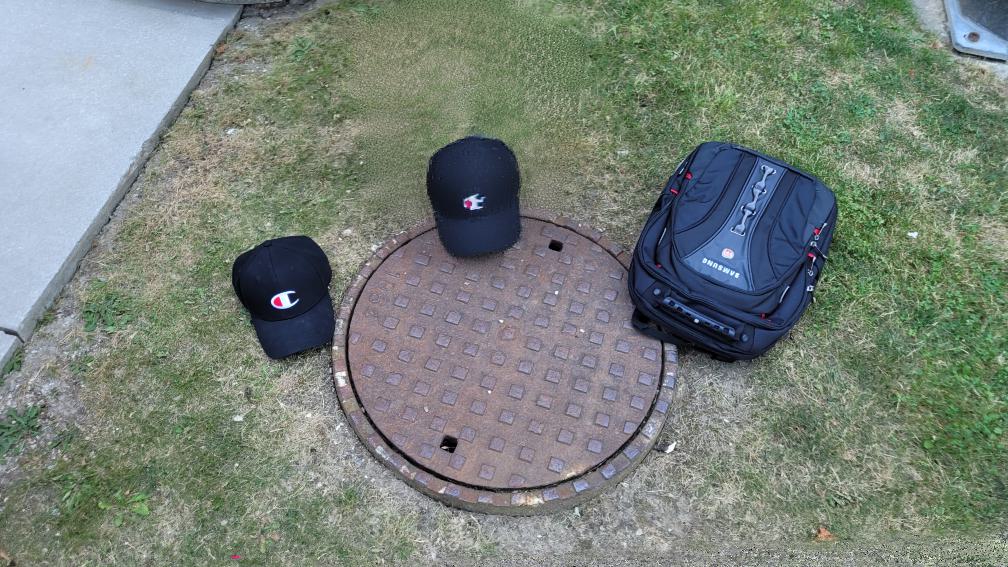} \\ [-0.3em]
        \midrule
        \parbox[t]{1.6em}{\rotatebox[origin=l]{90}{\scalebox{.5}{\hspace{1.5em}\makecell{Ours (Final)}}}} &
        \includegraphics[width=.24\linewidth,trim={8.7cm 6cm 9cm 4cm},clip]{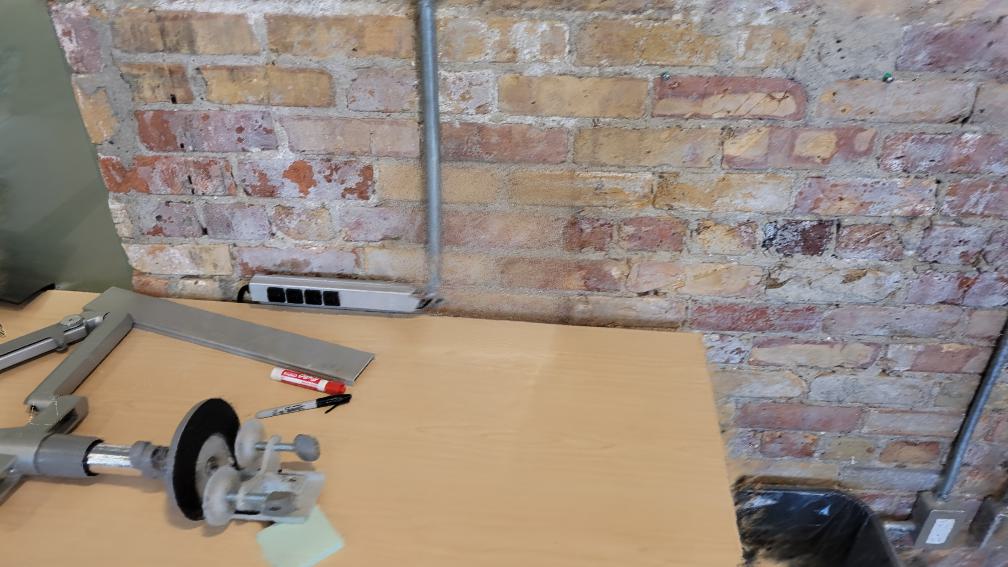} &
        \includegraphics[width=.24\linewidth,trim={8.7cm 6cm 9cm 4cm},clip]{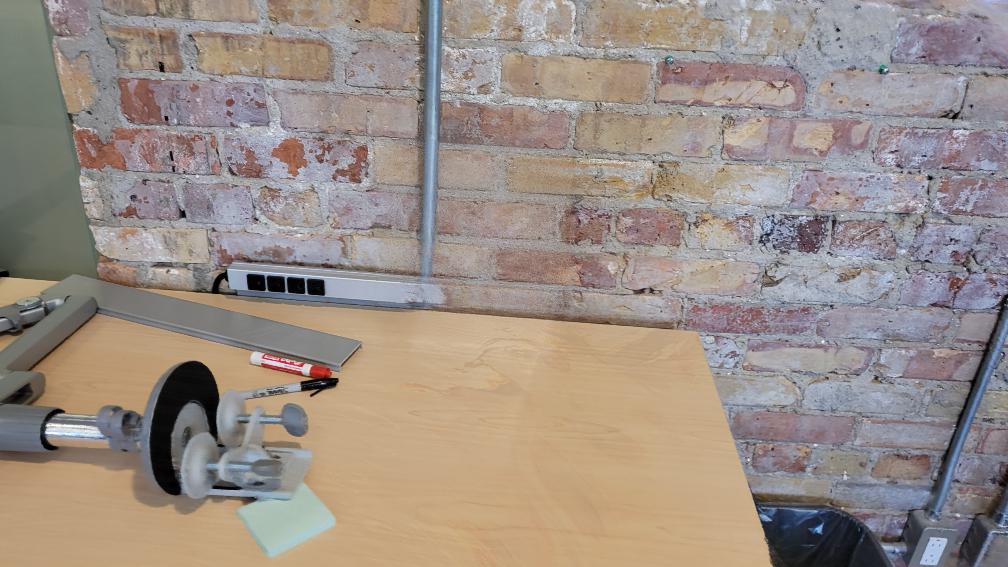} &
        \includegraphics[width=.24\linewidth,trim={5.5cm 8cm 8.7cm 0cm},clip]{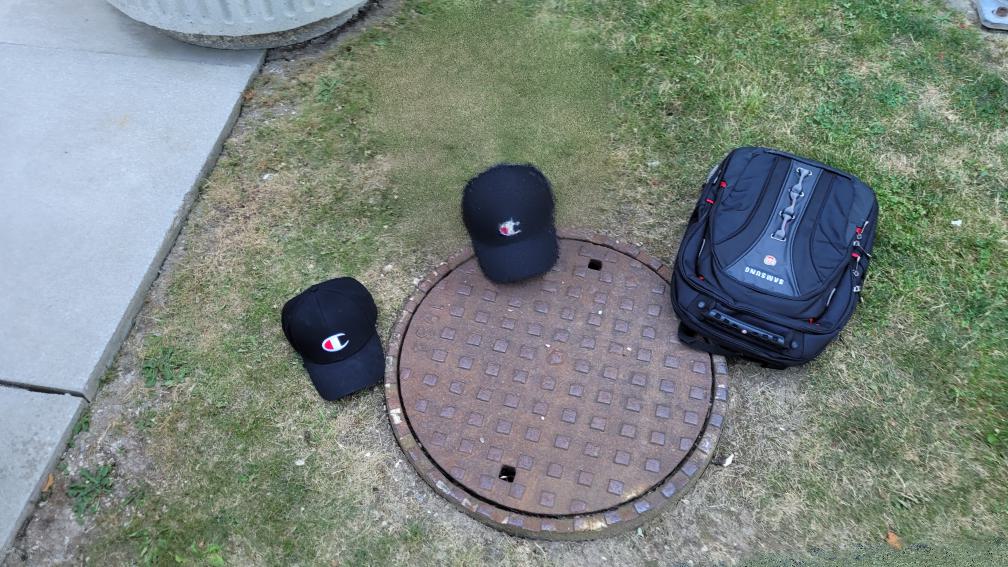} &
        \includegraphics[width=.24\linewidth,trim={7.5cm 8cm 6.7cm 0cm},clip]{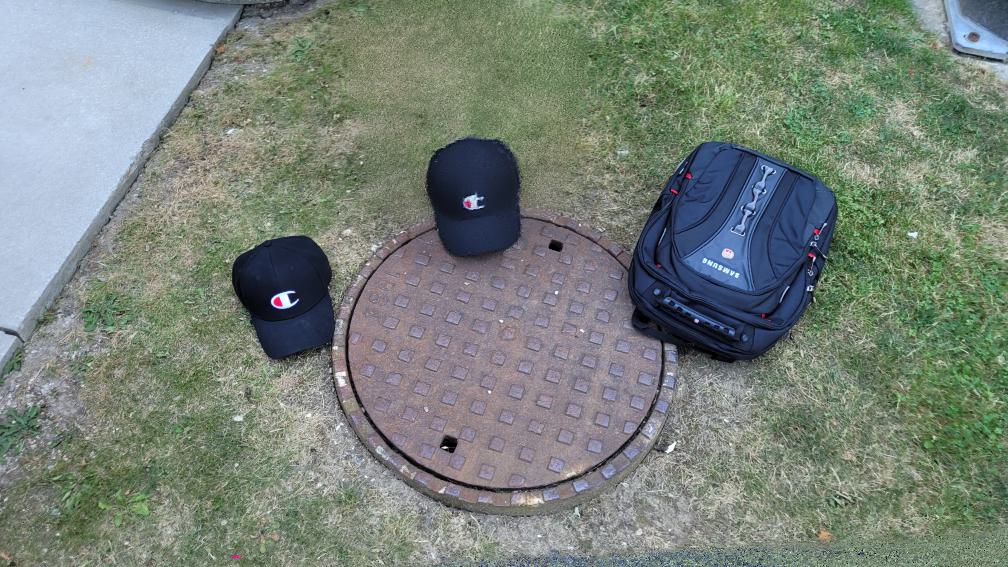} \\ [-0.3em]
        %
        %
        \bottomrule
    \end{tabular}
    \vspace{-.5em}
    \caption{\textbf{Qualitative ablation study.} In this study, we use the SPIn-NeRF dataset~\cite{spinnerf} to validate the effectiveness of masked adversarial loss, feature matching loss, per-scene customization, and if the commonly-used pixel-based (I-Recon) and LPIPS are needed.
    }
    \label{fig:ablation}
    \vspace{-1em}
\end{figure}

\begin{table}[t]
\centering
\setlength{\tabcolsep}{6pt}
\caption{\textbf{Quantitative ablation study.} We use the SPIn-NeRF dataset~\cite{spinnerf} to validate the effectiveness of masked adversarial loss, feature matching loss, per-scene customization, and if the common-used pixel-level (i.e., I-Recon) and LPIPS loss functions are required.
We highlight the performance which is worse than our complete approach {\color{red}red}, and mark the better one in {\color{green}green}.
}
\label{tab:ablation}
\scriptsize
\begin{tabular}{@{}lcccc@{}}
\toprule
Methods     & LPIPS ($\downarrow$) & M-LPIPS ($\downarrow$) & FID ($\downarrow$) & KID ($\downarrow$) \\ 
\midrule
Ours $-$ Adv $+$ I-Recon       & \degrade{0.6623} & \degrade{0.5236} & \degrade{305.60} & \degrade{0.1177} \\
Ours $-$ Adv $+$ LPIPS         & \improve{0.4231} & \improve{0.3147} & \degrade{192.86} & \degrade{0.0447} \\
Ours $-$ Adv $+$ I-Recon $+$ LPIPS & \degrade{0.4359} & \improve{0.3172} & \degrade{199.10} & \degrade{0.0458} \\
\midrule
Ours $+$ I-Recon               & \degrade{0.5106} & \degrade{0.3730} & \degrade{256.82} & \degrade{0.0827} \\
Ours $+$ LPIPS                 & \improve{0.4130} & \improve{0.3130} & \degrade{185.79} & \degrade{0.0419} \\
\midrule
Ours $-$ Per-Scene Finetune    & \degrade{0.4894} & \degrade{0.3862} & \degrade{224.29} & \degrade{0.0596} \\
Ours $-$ Feature Matching      & \degrade{0.4382} & \degrade{0.4002} & \degrade{232.28} & \degrade{0.0716} \\
Ours $-$ Adv Masking           & \degrade{0.4367} & \degrade{0.3358} & \degrade{196.47} & \degrade{0.0472} \\
\midrule
Ours                           & 0.4345           & 0.3344           & 183.25           & 0.0397  \\ 
\bottomrule
\end{tabular}
\end{table}

\vspace{\subsecmargin}
\subsection{Ablation Study}
\vspace{\subsecmargin}

We conduct a detailed ablation on the SPIn-NeRF dataset, which has a physically correct ground truth that measures the performance more reliably.
Our ablation study is conducted in three major sections.

We first show that removing our adversarial loss leads to worse visual quality in FID and KID measures, and such a performance gap cannot be closed by any combination of pixel reconstruction or LPIPS perceptual losses.
Note that, as mentioned in Figure~\ref{fig:lpips-problem}, the gain in the LPIPS score maintains high stochasticity and does not reflect the actual perceptual quality. 
Furthermore, it is expected that a method optimized toward the LPIPS networks should yield a more favorable LPIPS score, but the FID and KID scores indicate such a performance gain does not convert to a better visual quality.
The qualitative samples in Figure~\ref{fig:ablation} also show that the inpainted results from these methods maintain a much blurry appearance compared to our final method, but the LPIPS measurement is unable to reflect such an apparent blurriness.

Second, we ablate that neither pixel nor LPIPS losses can provide better visual quality when combined with our adversarial-based training scheme.
In particular, we found combining the pixel loss and our adversarial scheme causes instability that often leads to gradient explosion and fails the whole experiment.
On the other hand, as we discussed in the introduction and Figure~\ref{fig:discontinuity}, the LDM model creates discontinuities between the inpainting region and the real pixels.
The LPIPS objective, which has the receptive field crossing the discontinuous border, encourages to keeping a rather sharp border between the reconstruction and inpainting area, and even propagates the faulty gradients caused by the discontinuity into neighboring areas.
Such behavior leads to poorer geometry near the border of the inpainting mask.
In Figure~\ref{fig:ablation}, notice the ``Ours $+$ LPIPS'' method not only has a clear discontinuity on the brick wall and the grass area, but also has a poorer geometry on the iron tube and the baseball cap.
Note that the white artifacts on the table of the brick wall scene are irrelevant reflections of the foreground object introduced from other views, which is consistent and shared among all experimental results.

The third and the last part of the ablation shows the individual performance gain from each of our proposed components. 
In Table~\ref{tab:ablation}, removing any of the components leads to a consistent and significant performance loss in all measures.
In Figure~\ref{fig:ablation}, we show that the behavior of each proposed component is consistent with our motivation.
Removing the per-scene finetuning introduces random objects and creates obvious visual artifacts.
Removing the feature matching loss simply unstablizes the adversarial loss and creates obvious visual artifacts and wrong geometry.
Training the adversarial loss without our masked adversarial scheme encourages the discriminator to keep the discontinuity between the real and inpainted region, and leads to worse geometry near the inpainting border.
For instance, the continuity of the iron tube in the brick wall scene and the sharpness of the tree leaves in the garden scene are significantly impacted.


\vspace{\secmargin}
\section{Conclusions and Discussions}
\vspace{\secmargin}

In this work, we improve the NeRF inpainting performance by harnessing the latent diffusion model and solving optimization issues with a masked adversarial training scheme.
We justify the significance of each proposed component via careful comparisons against state-of-the-art baselines and rigorous ablation studies.
%

\noindent \textbf{Potential negative impact.}
NeRF inpainting with generative priors is closely related to generative inpainting, where certain frameworks aim to insert hallucinated content into a NeRF reconstruction. 
Such an application could lead to manipulating false information or creating factually wrong re-created renderings.
Although we do not focus on such an application, which requires extra effort to harmonize and shadow-cast the inserted objects, our techniques could be utilized by these methods.

\noindent \textbf{Limitations.}
Our framework involves an adversarial objective. Despite recent generative adversarial networks literature advancements, the framework's performance remains highly stochastic.
Also, it may not work well with few-shot NeRF reconstructions due to limited training data, or application scenarios with excessively large inpainting masks.
Our framework significantly improves NeRF's convergence by making the diffusion model generate consistent results across frames. 
However, our approach does not eliminate the microtextural variation caused by the stochastic denoising process. Therefore, the inpainted texture remains visually more blurry than the real-world textures.

\bibliographystyle{splncs04}
\bibliography{main}

\clearpage

{\centering\textbf{\Large Supplementary Material}}
\appendix
\begin{appendices}
\section{More Visual Results}

Please find more visual results and video renderings on our project page: {\url{https://hubert0527.github.io/MALD-NeRF}}.
%

\section{Object Insertion}

\begin{figure}[h]
    \centering
    \vspace{-1.2em}
    \includegraphics[width=.24\linewidth,trim={0 1.5cm 0 0},clip]{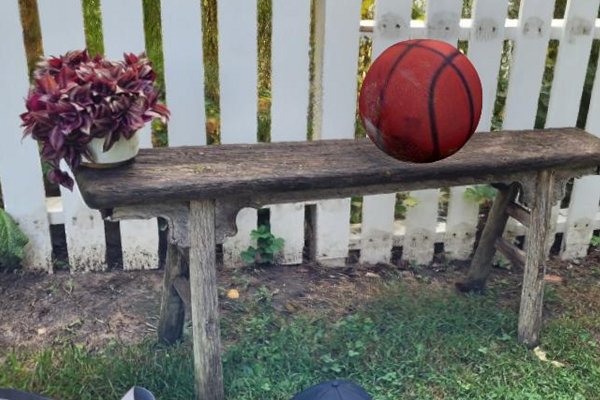} 
    \hfill
    \includegraphics[width=.24\linewidth,trim={0 1.5cm 0 0},clip]{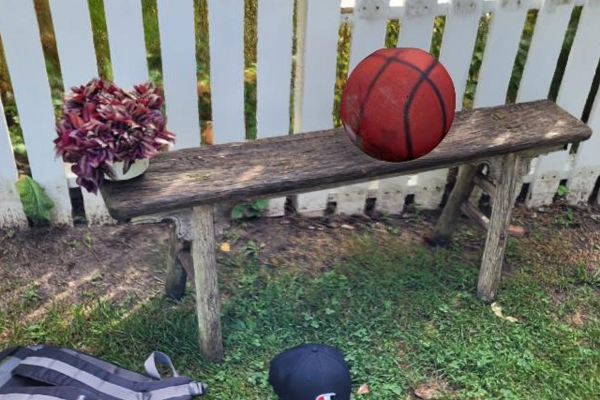}
    \hfill
    \includegraphics[width=.24\linewidth,trim={0 1.5cm 0 0},clip]{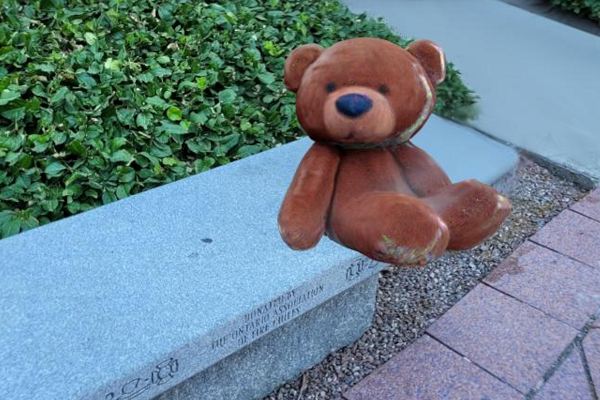}
    \hfill
    \includegraphics[width=.24\linewidth,trim={0 1.5cm 0 0},clip]{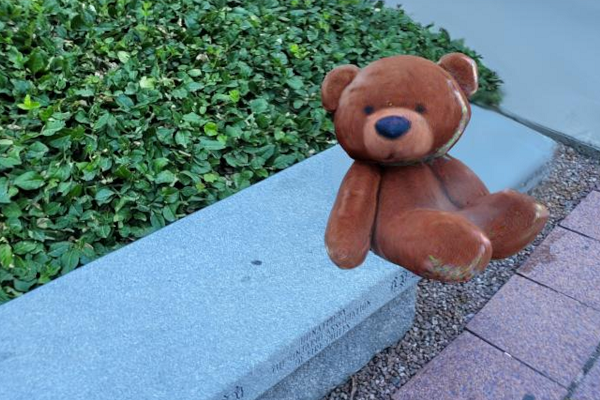}
    \vspace{-1em}
    \caption{Insert synthesized objects with depth blending.}
    \label{fig:insert}
    \vspace{-1.2em}
\end{figure}

A few prior works~\cite{mirzaei2023reference,weber2023nerfiller} also demonstrates applications of inpainting objects in NeRF with text prompts.
However, most of these results remain preliminary without complicated object-scene interaction.
In such cases, synthetic 3D objects can be inserted with depth blend afterward and later baked into NeRF with optimization.
Meanwhile, directly inpainting objects with the input masks is not effective as the masks strongly constrain the object shape.
In Figure~\ref{fig:insert}, similar to GaussianEditor~\cite{GaussianEditor}, we show that our framework can achieve object inpainting by first inpainting the background then followed by inserting generated objects.

\section{Training Robustness}
\setlength\intextsep{0pt}
\setlength{\columnsep}{5pt}
\begin{wraptable}[4]{r}{0.5\linewidth}
\centering
\renewcommand{\arraystretch}{0.5}
\setlength{\aboverulesep}{1.5pt}
\setlength{\belowrulesep}{1.5pt}
\setlength{\tabcolsep}{2pt}
\begin{tabular}{l c c c c}
    \toprule
    Trial & LPIPS & M-LPIPS & FID & KID \\
    \midrule
    A & 0.4147 & 0.3056 & 195.47 & 0.0466 \\
    B & 0.4138 & 0.3055 & 194.59 & 0.0456 \\
    C & 0.4142 & 0.3058 & 191.91 & 0.0486 \\
    \bottomrule
\end{tabular}
\end{wraptable}
%
Despite we adopt adversarial loss in our optimization objectives, our algorithm is stable with R$_1$ regularizer~\cite{mescheder2018training} and gradient penalty~\cite{gulrajani2017improved}.
We show the robustness by reporting the performance of three individual training trials of all scenes from SPIn-NeRF dataset.
To save computation with the large amount of experiments, we use 1 GPU instead of 8 GPUs per experiment.

\newlength{\oldintextsep}
\setlength{\oldintextsep}{\intextsep}
\setlength\intextsep{0pt}
\setlength{\columnsep}{3pt}
\begin{wrapfigure}[7]{r}{0.32\linewidth}
    \centering
    \includegraphics[width=\linewidth]{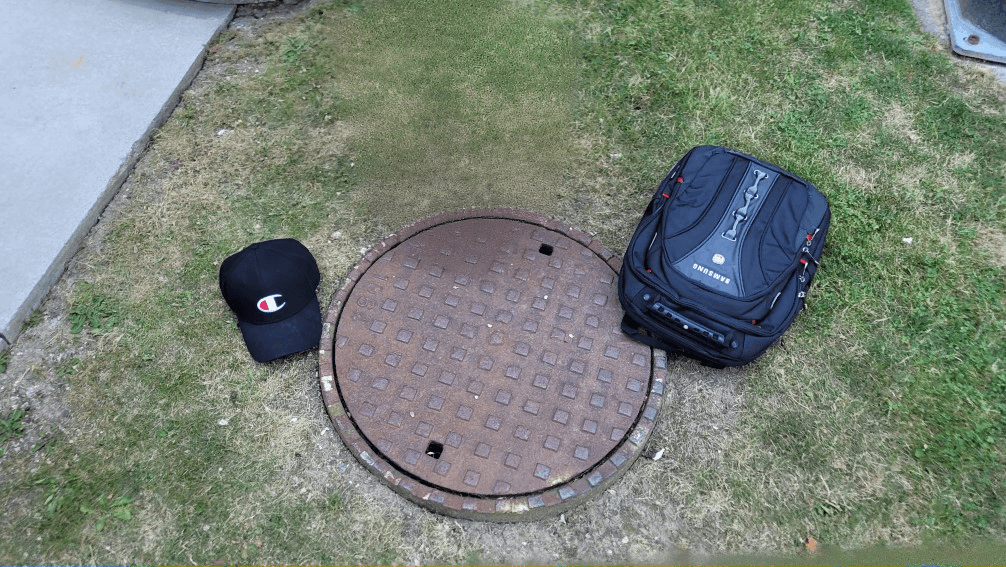}
    \vspace{-1.5em}
\end{wrapfigure}
\section{Further Customization}
During the per-scene customization, due to repetitively using the same set of images as the finetuning data distribution, the model can sometimes memorize certain scene-dependent content.
The behavior can lead to inpainting objects in the inpainting region, such as the baseball cap inpainted in Figure 5 of the main paper.
Such a behavior can easily remedy by masking the unwanted objects from the scene-dependent customization.
We show such a result on the right.

\section{Evaluate Geometric Consistency with Optical Flow-based Metric}
Following \cite{tseng2023consistent}, we evaluate the cross-frame consistency with a flow warping score. We treat the image set of each scene as an image sequence, then calculate optical flow between an image pair $I_t$ and $I_{t+1}$ with RAFT~\cite{teed2020raft}, finally, obtain $\tilde{I}_t$ by warping $I_{t+1}$ with the optical flow. 
We compute the consistency with $M_{t \rightarrow t+1} \cdot \| I_t - \tilde{I}_t \|_1$, where $M_{t \rightarrow t+1}$ is the disocclusion mask from time $t$ to $t+1$.
We found such a metric highly depends on the flow quality predicted by RAFT, and heavily favors blurry inpainting results.
In practice, we found the blurry results from all baselines outperforms the consistency scores from the real images.

\section{Implementation Details}

\vspace{\paraskip}
\noindent \textbf{NeRF.}
We use a self-implemented NeRF framework similar to ZipNeRF~\cite{barron2023zip} that uses hash-based~\cite{muller2022instant} positional encoding along with multiple MLPs to predict the final density and RGB quantities.
A scene contraction is applied to the NeRF~\cite{barron2022mip} as all the scenes we experimented on are unbounded scenes.
We use two proposal networks to perform importance sampling, followed by the main network. 
The network designs are similar to the Nerfacto implemented in the Nerfstudio~\cite{nerfstudio}.

\vspace{\paraskip}
\noindent \textbf{Hyperparameters.}
For all experiments, we train the networks with 8 V100 GPUs for 30{,}000 iterations at a ray batch size of 16{,}384 using distributed data-parallel. 
The choice of batch size is constrained by the amount of GPU VRAM after loading the NeRF and other deep image priors, such as the latent diffusion model network for generative inpainting and the ZoeDepth model (NK version) for the depth loss.
All these deep image priors are inferenced without calculating gradients to reduce VRAM usage, and inference at a batch size of 1.

We use two separate optimizers for NeRF reconstruction and adversarial learning.
The NeRF reconstruction uses an Adam optimizer with a learning rate decay from $0.01$ to $0.0001$, while adversarial learning uses an Adam optimizer with a learning rate $0.0001$ throughout the training.
Different from GANeRF~\cite{roessle2023ganerf}, we found using RMSProp makes the training unstable.
For the adversarial learning, we use a discriminator architecture similar to StyleGAN2~\cite{stylegan2}.
We train the discriminator with 64$\times$64 patches.
We importance-sample 256$\times$256 image patches based on the number of inpainting pixels within the patch, then slice the image patch into the discriminator training patch and train the discriminator at a batch size of 16.
For the importance sampling strategy, we first exclude patches with insufficient inpainting pixels (we empirically set the threshold to 50\%). 
Assume that each patch index $i$ contains $d_i$ inpainting pixels, we assign a probability $p_i = d_i / \sum_j d_j$ while sampling the patches for training.

As mentioned in Eq~$4$, Eq~$5$, and Eq~$6$ of the main paper, our networks are being trained with various loss terms. 
We balance the loss terms with $\lambda_\text{pix}=1$, $\lambda_\text{inter}=3$, $\lambda_\text{distort}=0.002$, $\lambda_\text{decay}=0.1$, $\lambda_\text{adv}=1$, $\lambda_\text{fm}=1$ and $\lambda_\text{GP}=15$.

\vspace{\paraskip}
\noindent \textbf{Iterative Dataset Update.}
We infer the latent diffusion model with a DDIM scheduler for 20 steps.
During the iterative dataset update, we synchronize the random sampled image IDs across GPUs to ensure there is no overlap among GPUs, then update the 8 distinctly sampled images in the dataset with partial DDIM.
For the partial DDIM, we first hard-blend the rendered pixels in the inpainting mask region with the real pixels outside the inpainting region into a 512$\times$512 image, encode into the latent space with the auto-encoder of the latent diffusion model, then add the noise level at timestep $t$ based on the current training progress and the HiFA scheduling.
Therefore, as the training progresses, the final inpainted images will gradually converge to the current NeRF rendering results due to low noise levels.
Since we update 8 images in each dataset update step, we set the frequency of iterative dataset update to one dataset update every 80 NeRF training steps, which is 8 times less frequent compared to InstructNeRF2NeRF~\cite{haque2023instruct}.
The whole training approximately takes 16 hours on 8 V100 GPUs.
\end{appendices}

\clearpage

%
%
\end{document}